\documentclass[11pt,a4paper]{article}

\usepackage[nopatch=footnote]{microtype}

\usepackage{fontspec}
\usepackage{polyglossia}
\setdefaultlanguage{english}
\setotherlanguage{bengali}
\newfontfamily\bengalifont[
  Script=Bengali,
  Path=fonts/,
  Extension=.ttf,
  UprightFont=*-Regular,
  BoldFont=*-Bold
]{NotoSerifBengali}

\usepackage[top=2.5cm,bottom=2.5cm,left=2.5cm,right=2.5cm]{geometry}

\usepackage{booktabs}
\usepackage{tabularx}
\usepackage{multirow}
\usepackage{array}
\usepackage{makecell}
\usepackage{adjustbox}

\usepackage{graphicx} 
\usepackage{float}
\usepackage{caption}
\usepackage{subcaption}

\usepackage{pgfplots}
\pgfplotsset{compat=1.18}
\usepackage{pgfplotstable}
\usepackage{tikz}
\usetikzlibrary{patterns,decorations.pathreplacing,positioning}

\usepackage{amsmath}
\usepackage{amssymb}

\usepackage{algorithm}
\usepackage{algpseudocode}
\algrenewcommand\algorithmicrequire{\textbf{Input:}}
\algrenewcommand\algorithmicensure{\textbf{Output:}}

\usepackage{hyperref}
\hypersetup{
    colorlinks=true,
    linkcolor=blue!60!black,
    citecolor=blue!60!black,
    urlcolor=blue!60!black
}

\usepackage{natbib}
\usepackage{xcolor}
\usepackage{soul}
\usepackage{enumitem}
\usepackage{setspace}
\definecolor{bsmddcol}{RGB}{115,114,108}
\definecolor{ytcol}{RGB}{235,104,52}
\definecolor{daptcol}{RGB}{27,175,122}
\definecolor{multicol}{RGB}{42,120,214}
\definecolor{goldcol}{RGB}{115,114,108}
\definecolor{combinedcol}{RGB}{27,175,122}

\pgfplotsset{
  every axis/.style={
    grid=major,
    grid style={dashed,gray!25},
    tick label style={font=\small},
    label style={font=\small},
    legend style={font=\small},
  }
}

\newcommand{\cryingemoji}{%
  \tikz[baseline=-0.5ex,scale=0.18]{
    \fill[yellow!88!orange] (0,0) circle (1);
    \draw[black,line width=0.55pt] (-0.61,0.58) .. controls (-0.45,0.78) and (-0.27,0.75) .. (-0.16,0.56);
    \draw[black,line width=0.55pt] (0.16,0.56) .. controls (0.27,0.75) and (0.45,0.78) .. (0.61,0.58);
    \fill[black] (-0.36,0.27) circle (0.085);
    \fill[black] (0.36,0.27) circle (0.085);
    \fill[cyan!75!blue] (-0.46,0.14) .. controls (-0.61,-0.08) and (-0.59,-0.34) .. (-0.44,-0.48) .. controls (-0.29,-0.36) and (-0.32,-0.11) .. cycle;
    \fill[cyan!75!blue] (0.46,0.14) .. controls (0.61,-0.08) and (0.59,-0.34) .. (0.44,-0.48) .. controls (0.29,-0.36) and (0.32,-0.11) .. cycle;
    \fill[black] (0,-0.39) ellipse [x radius=0.17, y radius=0.23];
  }
}

\newcommand{\sadfaceemoji}{%
  \tikz[baseline=-0.5ex,scale=0.18]{%
    \fill[yellow!88!orange] (0,0) circle (1);
    \fill[black] (-0.36,0.29) circle (0.08);
    \fill[black] (0.36,0.29) circle (0.08);
    \draw[black,line width=0.65pt] (-0.42,-0.13) .. controls (-0.22,0.18) and (0.22,0.18) .. (0.42,-0.13);
    \fill[cyan!75!blue] (0.48,0.12) .. controls (0.35,-0.08) and (0.37,-0.28) .. (0.49,-0.36) .. controls (0.61,-0.25) and (0.61,-0.06) .. cycle;
  }
}

\newcommand{\brokenheartemoji}{%
  \tikz[baseline=-0.5ex,scale=0.18]{
    \fill[red!82!magenta] (0,0.76) .. controls (-0.25,1.12) and (-0.84,1.08) .. (-0.88,0.50) .. controls (-0.92,0.12) and (-0.46,-0.42) .. (0,-0.82) .. controls (0.46,-0.42) and (0.92,0.12) .. (0.88,0.50) .. controls (0.84,1.08) and (0.25,1.12) .. (0,0.76) -- cycle;
    \draw[white,line width=0.75pt] (0,0.82) -- (-0.16,0.28) -- (0.13,0.03) -- (-0.15,-0.30) -- (0,-0.78);
  }
}

\newcommand{\smileemoji}{%
  \tikz[baseline=-0.5ex,scale=0.18]{
    \fill[yellow!88!orange] (0,0) circle (1);
    \fill[black] (-0.36,0.29) circle (0.08);
    \fill[black] (0.36,0.29) circle (0.08);
    \draw[black,line width=0.65pt] (-0.45,-0.15) .. controls (-0.25,-0.58) and (0.25,-0.58) .. (0.45,-0.15);
  }
}

\newcommand{\pleadingemoji}{%
  \tikz[baseline=-0.5ex,scale=0.18]{
    \fill[yellow!88!orange] (0,0) circle (1);
    \draw[black,line width=0.55pt] (-0.62,0.57) .. controls (-0.43,0.76) and (-0.26,0.73) .. (-0.14,0.57);
    \draw[black,line width=0.55pt] (0.14,0.57) .. controls (0.26,0.73) and (0.43,0.76) .. (0.62,0.57);
    \fill[blue!68!cyan] (-0.36,0.27) ellipse [x radius=0.17, y radius=0.20];
    \fill[blue!68!cyan] (0.36,0.27) ellipse [x radius=0.17, y radius=0.20];
    \fill[white] (-0.40,0.34) circle (0.045);
    \fill[white] (0.32,0.34) circle (0.045);
    \fill[black] (0,-0.40) ellipse [x radius=0.10, y radius=0.14];
  }
}

\newcommand{\pensiveemoji}{%
  \tikz[baseline=-0.5ex,scale=0.18]{%
    \fill[yellow!88!orange] (0,0) circle (1);
    \draw[black,line width=0.55pt] (-0.62,0.57) .. controls (-0.42,0.75) and (-0.24,0.70) .. (-0.12,0.54);
    \draw[black,line width=0.55pt] (0.15,0.64) .. controls (0.36,0.77) and (0.55,0.69) .. (0.65,0.48);
    \fill[black] (-0.34,0.23) ellipse [x radius=0.09, y radius=0.12];
    \fill[black] (0.34,0.23) ellipse [x radius=0.09, y radius=0.12];
    \draw[black,line width=0.6pt] (-0.18,-0.39) .. controls (0,-0.30) and (0.16,-0.30) .. (0.24,-0.40);
  }
}

\newcommand{\wearyemoji}{%
  \tikz[baseline=-0.5ex,scale=0.18]{%
    \fill[yellow!88!orange] (0,0) circle (1);
    \draw[black,line width=0.8pt] (-0.52,0.31) .. controls (-0.37,0.12) and (-0.20,0.12) .. (-0.08,0.31);
    \draw[black,line width=0.8pt] (0.08,0.31) .. controls (0.20,0.12) and (0.37,0.12) .. (0.52,0.31);
    \fill[black] (0,-0.39) ellipse [x radius=0.16, y radius=0.22];
  }
}

\newcommand{\skullemoji}{%
  \tikz[baseline=-0.5ex,scale=0.18]{%
    \fill[white] (0,0.12) ellipse [x radius=0.62, y radius=0.76];
    \draw[black,line width=0.6pt] (0,0.12) ellipse [x radius=0.62, y radius=0.76];
    \fill[black] (-0.25,0.22) ellipse [x radius=0.15, y radius=0.20];
    \fill[black] (0.25,0.22) ellipse [x radius=0.15, y radius=0.20];
    \fill[black] (0,-0.08) -- (-0.10,-0.27) -- (0.10,-0.27) -- cycle;
    \draw[black,line width=0.55pt] (-0.36,-0.43) -- (0.36,-0.43);
    \draw[black,line width=0.45pt] (-0.20,-0.43) -- (-0.20,-0.61);
    \draw[black,line width=0.45pt] (0,-0.43) -- (0,-0.64);
    \draw[black,line width=0.45pt] (0.20,-0.43) -- (0.20,-0.61);
  }
}

\newcommand{\skullcrossbonesemoji}{%
  \tikz[baseline=-0.5ex,scale=0.18]{%
    \draw[gray!65,line width=1.4pt,line cap=round] (-0.72,-0.75) -- (0.72,0.72);
    \draw[gray!65,line width=1.4pt,line cap=round] (-0.72,0.72) -- (0.72,-0.75);
    \fill[white] (-0.72,-0.75) circle (0.10);
    \fill[white] (0.72,0.72) circle (0.10);
    \fill[white] (-0.72,0.72) circle (0.10);
    \fill[white] (0.72,-0.75) circle (0.10);
    \draw[black,line width=0.45pt] (-0.72,-0.75) circle (0.10);
    \draw[black,line width=0.45pt] (0.72,0.72) circle (0.10);
    \draw[black,line width=0.45pt] (-0.72,0.72) circle (0.10);
    \draw[black,line width=0.45pt] (0.72,-0.75) circle (0.10);
    \fill[white] (0,0.12) ellipse [x radius=0.62, y radius=0.76];
    \draw[black,line width=0.6pt] (0,0.12) ellipse [x radius=0.62, y radius=0.76];
    \fill[black] (-0.25,0.22) ellipse [x radius=0.15, y radius=0.20];
    \fill[black] (0.25,0.22) ellipse [x radius=0.15, y radius=0.20];
    \fill[black] (0,-0.08) -- (-0.10,-0.27) -- (0.10,-0.27) -- cycle;
    \draw[black,line width=0.55pt] (-0.36,-0.43) -- (0.36,-0.43);
    \draw[black,line width=0.45pt] (-0.20,-0.43) -- (-0.20,-0.61);
    \draw[black,line width=0.45pt] (0,-0.43) -- (0,-0.64);
    \draw[black,line width=0.45pt] (0.20,-0.43) -- (0.20,-0.61);
  }
}

\newcommand{\disappointedemoji}{%
  \tikz[baseline=-0.5ex,scale=0.18]{%
    \fill[yellow!88!orange] (0,0) circle (1);
    \fill[black] (-0.36,0.27) circle (0.08);
    \fill[black] (0.36,0.27) circle (0.08);
    \draw[black,line width=0.65pt] (-0.42,-0.16) .. controls (-0.22,0.16) and (0.22,0.16) .. (0.42,-0.16);
  }
}

\newcommand{\worriedemoji}{%
  \tikz[baseline=-0.5ex,scale=0.18]{%
    \fill[yellow!88!orange] (0,0) circle (1);
    \draw[black,line width=0.55pt] (-0.58,0.70) .. controls (-0.38,0.53) and (-0.22,0.53) .. (-0.10,0.65);
    \draw[black,line width=0.55pt] (0.10,0.65) .. controls (0.22,0.53) and (0.38,0.53) .. (0.58,0.70);
    \fill[black] (-0.35,0.27) ellipse [x radius=0.08, y radius=0.12];
    \fill[black] (0.35,0.27) ellipse [x radius=0.08, y radius=0.12];
    \fill[black] (0,-0.39) ellipse [x radius=0.11, y radius=0.16];
  }
}

\newcommand{\laughtearsemoji}{%
  \tikz[baseline=-0.5ex,scale=0.18]{%
    \fill[yellow!88!orange] (0,0) circle (1);
    \draw[black,line width=0.65pt] (-0.55,0.32) .. controls (-0.40,0.07) and (-0.22,0.07) .. (-0.10,0.31);
    \draw[black,line width=0.65pt] (0.10,0.31) .. controls (0.22,0.07) and (0.40,0.07) .. (0.55,0.32);
    \fill[red!70!magenta] (0,-0.34) ellipse [x radius=0.24, y radius=0.28];
    \fill[cyan!75!blue] (-0.62,0.15) .. controls (-0.77,-0.08) and (-0.73,-0.31) .. (-0.60,-0.42) .. controls (-0.47,-0.30) and (-0.50,-0.05) .. cycle;
    \fill[cyan!75!blue] (0.62,0.15) .. controls (0.77,-0.08) and (0.73,-0.31) .. (0.60,-0.42) .. controls (0.47,-0.30) and (0.50,-0.05) .. cycle;
  }
}

\newcommand{\prayeremoji}{%
  \tikz[baseline=-0.5ex,scale=0.18]{%
    \fill[blue!70] (-0.43,-0.78) -- (-0.08,-0.58) -- (0,-0.78) -- (0.08,-0.58) -- (0.43,-0.78) -- (0.35,-1) -- (-0.35,-1) -- cycle;
    \fill[orange!82!brown] (-0.08,-0.66) -- (-0.36,-0.39) -- (-0.42,0.50) .. controls (-0.43,0.72) and (-0.31,0.85) .. (-0.21,0.75) -- (-0.18,0.05) -- (-0.13,0.92) .. controls (-0.11,1.08) and (0.02,1.08) .. (0.03,0.91) -- (0.05,0.08) -- (0.11,0.85) .. controls (0.13,1.00) and (0.25,0.99) .. (0.26,0.82) -- (0.25,0.05) -- (0.34,0.65) .. controls (0.38,0.82) and (0.50,0.78) .. (0.49,0.59) -- (0.46,-0.28) -- (0.14,-0.69) -- cycle;
    \draw[black,line width=0.45pt] (-0.08,-0.66) -- (-0.36,-0.39) -- (-0.42,0.50) .. controls (-0.43,0.72) and (-0.31,0.85) .. (-0.21,0.75) -- (-0.18,0.05) -- (-0.13,0.92) .. controls (-0.11,1.08) and (0.02,1.08) .. (0.03,0.91) -- (0.05,0.08) -- (0.11,0.85) .. controls (0.13,1.00) and (0.25,0.99) .. (0.26,0.82) -- (0.25,0.05) -- (0.34,0.65) .. controls (0.38,0.82) and (0.50,0.78) .. (0.49,0.59) -- (0.46,-0.28) -- (0.14,-0.69) -- cycle;
    \draw[black,line width=0.45pt] (0.02,0.75) -- (0.05,0.10);
  }
}
\begin{document}

\title{\textbf{BanglishDepYT: Depression Detection in Native Bengali
Social Media with Emotion-Preserving Preprocessing and
Domain-Adaptive Pretraining}}

\author{%
  \begin{minipage}[t]{0.31\textwidth}
    \centering
    \textbf{Ayon Roy}\thanks{Corresponding author.}\\[3pt]
    \small Cyber Physical and Social Systems\\
    Université Jean Monnet, France\\
    \url{ayon.roy@etu.univ-st-etienne.fr}
  \end{minipage}\hfill
  \begin{minipage}[t]{0.31\textwidth}
    \centering
    \textbf{Sadman Sharif}\\[3pt]
    \small Lecturer of Innovation \& Entrepreneurship\\
    Daffodil International University, Bangladesh\\
    \url{sharif.die@diu.edu.bd}
  \end{minipage}\hfill
  \begin{minipage}[t]{0.31\textwidth}
    \centering
    \textbf{Syed Shams Islam}\\[3pt]
    \small Lecturer of Software Engineering\\
    Daffodil International University, Bangladesh\\
    \url{shams.swe@diu.edu.bd}
  \end{minipage}%
}

\date{}
\maketitle


\begin{abstract}
Depression affects millions of Bengali speakers across South Asia, yet automated detection systems for this population have relied almost entirely on datasets translated from English---datasets that structurally cannot contain the emoticons, romanized Bengali words, and expressive elongated spelling that characterize how people actually write when they are in distress. We introduce BanglishDepYT, a Bengali depression dataset collected natively from YouTube mental health comment sections, comprising 191,000 unlabeled comments and 1,000 samples verified through dual bilingual annotation. Alongside the corpus, we propose EPH (Emotion-Preserving Hybrid preprocessing), a text normalization algorithm that converts emoticons to semantic tags, preserves Banglish code-mixed tokens, protects negation words, and normalizes emotional elongation rather than erasing it. Preprocessing choice can substantially affect classification performance on informal social media text, although the effect depends on the model: standard cleaning reduces the Banglish Survival Rate of native Bengali social media to zero, and on a fused native-text corpus EPH retains more of the measured signal than standard preprocessing (CESS: 0.9996 versus 0.6694). On BanglishDepYT, EPH raises TF-IDF SVM F1 from 0.7062 to 0.8570 (+0.1508) and BanglaBERT F1 from 0.8355 to 0.8621, while BiLSTM and SahajBERT show no gain. We further investigate domain-adaptive pretraining (DAPT), continuing SahajBERT's pretraining on unlabeled comments from the target domain before fine-tuning. Across ten training seeds on a fixed 150-comment test set, DAPT-SahajBERT achieves the highest mean F1 (0.8161) and MCC (0.6272) among the evaluated transformers, outperforming the SahajBERT baseline (F1: 0.7805; paired $t$-test over seeds $p=0.0013$, $d_z=1.450$; higher F1 on 8 of 10 seeds) and performing comparably to MuRIL ($p=0.589$). At the level of individual test predictions the difference is not significant (McNemar $p=0.170$), so this result requires confirmation on a larger test set. LLM-assisted annotation with a locally deployed Qwen2.5-7B model achieves substantial agreement with human labels ($\kappa=0.611$), yielding 96,071 high-confidence pseudo-labels, of which 66,851 remain after preprocessing, gold-overlap removal and deduplication. Semi-supervised training on the combined dataset improves AUC from 0.8499 to 0.8731 ($p=0.044$), offering a way to scale annotation without proportional human effort. All code, models, and datasets are publicly released to support reproducible Bengali mental health NLP research.
\end{abstract}


\section{Introduction}
\label{sec:introduction}

Depression is one of the most serious mental health crises of our
time. The World Health Organization estimates that more than 280
million people worldwide live with depression \citep{who2023}, and
the burden falls particularly hard on South Asia. In Bangladesh
alone, an estimated 4.5 million people are affected
\citep{bangladesh2022}. Despite this scale, most people who need
help never receive it. There are not enough psychiatrists. Clinics
are too expensive. And in many communities, talking about mental
health is still considered shameful. People stay silent in person,
but they do not stay silent online.

Every day, millions of Bengali speakers go to YouTube, Facebook,
and other platforms to express what they cannot say out loud. They
write under mental health videos, under sad songs, under personal
story channels. They write in the language they actually use, which
is a mixture of Bengali script, English words typed phonetically,
emotional repetition, and emoji that carry entire sentences of
feeling. A comment like \textit{``ami r parchi na \cryingemoji
ekla lagche onek''}---I cannot take it anymore, I feel so
alone---carries something important. It carries it without formal
grammar, without clinical terminology, and without a doctor
present. If a system could read these signals reliably and early,
it might be possible to identify people who are struggling before
a crisis happens.

This is the promise of automated depression detection from social
media text. But there is a serious problem with how this research
has been done for Bengali.

Almost every existing Bengali depression dataset was built by
taking posts written in English, from Reddit communities like
\textit{r/depression} and from Twitter, and translating them into
Bengali \citep{chowdhury2024}. The resulting datasets look like
Bengali but do not sound like it. They contain no emoticons,
because translators had no reason to insert them. They contain no
Banglish, no romanized expressions like \textit{kosto lagche} or
\textit{mon kharap}, because translators wrote in standard script.
They contain no elongated words like \textbengali{আআআআ} that
signal emotional intensity through repetition. In short, these
datasets are missing exactly the features that make real Bengali
social media text emotionally expressive, and models trained on
them have never encountered the informal language they would face
in any real deployment.

This gap is not a minor technical detail. When we measure how much
standard preprocessing removes from native Bengali social media
text, we find that the Banglish Survival Rate drops to exactly
zero. Every single romanized word is deleted before the model sees
the text. The most widely used Bengali depression benchmark, BSMDD
\citep{chowdhury2024}, scores identically under raw, minimal, and
standard preprocessing conditions, revealing that it was released
in pre-processed form, making prior preprocessing comparisons on
that dataset uninformative. Models trained on translated,
pre-processed data and evaluated on the same kind of data look
strong on paper. They have simply never been tested on anything
that resembles real informal Bengali.

We address this gap with three contributions.

\noindent\textbf{First}, we introduce \textbf{BanglishDepYT},
a Bengali depression dataset collected natively from
YouTube. We collected 191{,}000 Bengali comments from mental
health related video discussions using 150 search queries across
13 thematic categories, applied quality filtering to produce a
clean corpus of 163{,}217 comments, and had two bilingual
annotators, both authors of this paper, label 1{,}000
samples through a discussion-based consensus process. Unlike
translated datasets, BanglishDepYT retains informal Bengali
expression: across the full clean corpus, 22.4\% of comments
include emoticons, 7.9\% include Banglish tokens, and 16.6\%
include elongated emotional phrasing.

\noindent\textbf{Second}, we propose \textbf{EPH}
(Emotion-Preserving Hybrid preprocessing), a text normalization
method designed specifically for informal Bengali social media.
EPH converts emoticons to semantic tags such as \texttt{[CRY]}
and \texttt{[BROKEN\_HEART]}, preserves Banglish tokens, retains
negation words like \textbengali{না} and \textbengali{নেই} which
are critical to the meaning of depression-related sentences, and
normalizes but does not erase emotional elongation. We introduce
the Composite Emotional Signal Score (CESS) as a diagnostic metric
combining information preservation, Banglish survival, and
emotional lexicon preservation to quantify how much emotional
signal any preprocessing condition retains. On the fused native
corpus, EPH achieves CESS of 0.9996 against 0.6694 for standard
preprocessing. On BanglishDepYT it produces a fifteen-point F1
improvement for a TF-IDF SVM and a smaller gain for BanglaBERT,
but no gain for BiLSTM or SahajBERT. The benefit of
emotion-preserving preprocessing is therefore model-dependent;
our results do not establish that preprocessing matters more
than model architecture in general.

\noindent\textbf{Third}, we demonstrate two complementary
strategies for improving performance on native Bengali text.
Domain-adaptive pretraining (DAPT) adapts SahajBERT to informal
Bengali by continuing its pretraining on the full 163{,}217
unlabeled clean comments before fine-tuning on the labeled set.
Across ten training seeds, \textbf{DAPT-SahajBERT} obtains the
highest mean F1 and MCC among the evaluated transformers and
outperforms the monolingual baseline (F1: 0.8161 versus 0.7805,
paired $t$-test over seeds $p = 0.0013$, $d_z = 1.450$; higher F1
on 8 of 10 seeds) while performing comparably to MuRIL, a
multilingual model trained on vastly larger corpora
($p = 0.589$). Because all comparisons share one 150-comment test
set, on which the prediction-level difference is not significant
(McNemar $p = 0.170$), we treat this as evidence that targeted
domain pretraining is a promising, inexpensive alternative to
large-scale multilingual pretraining, not as a settled result. LLM-assisted annotation using
a locally deployed Qwen2.5-7B model achieves substantial human
agreement (Cohen's $\kappa = 0.611$) and pseudo-labels 96{,}071
additional comments, of which 66{,}851 remain after
preprocessing, gold-overlap removal and deduplication.
Semi-supervised training on the combined
dataset raises AUC from 0.8499 to 0.8731 ($p = 0.044$),
providing a reproducible zero-cost annotation scaling pathway
for mental health NLP in data-scarce languages.

Together these contributions address a problem the field has
largely overlooked: the informal linguistic reality of Bengali
mental health discourse online is absent from the translated
benchmarks most commonly used for this task, and the one prior
native collection we are aware of \citep{ghosh2023} removed
emoticons and non-Bangla characters before modeling.

The rest of this paper is organized as follows.
Section~\ref{sec:related_work} reviews related work on mental
health NLP, Bengali language processing, preprocessing effects,
and domain adaptation.
Section~\ref{sec:dataset} describes BanglishDepYT and its
collection, annotation, and LLM-assisted pseudo-labeling
methodology.
Section~\ref{sec:eph} presents the EPH algorithm and the CESS
metric framework.
Section~\ref{sec:experimental_setup} details the experimental
setup.
Section~\ref{sec:results} reports results across all model
conditions.
Section~\ref{sec:stats} presents statistical analysis and error
analysis.
Section~8 discusses limitations and ethical considerations.
Section~9 concludes with a summary of findings and future
directions. All code, models, and datasets are publicly released
to support reproducible Bengali mental health NLP research.


\section{Related Work}
\label{sec:related_work}

\subsection{Depression Detection from Social Media Text}

Researchers have been trying to detect depression from the way
people write online for close to a decade. The earliest systems
were straightforward: count emotionally weighted words, feed them
into a Support Vector Machine, and see how well the classifier
separates depressive posts from healthy ones
\citep{coppersmith2015}. This worked to a degree on English Reddit
and Twitter data, but it collapsed whenever the vocabulary shifted
slightly, because the model had no understanding of context at all.
Deep learning changed this. LSTM and CNN-LSTM architectures could
at least follow word order and capture short-range dependencies,
which mattered for negation --- ``I am not okay'' and ``I am okay''
look identical to a bag-of-words model \citep{mumu2023,uddin2022}.

The real breakthrough for mental health NLP came with BERT and the
insight that domain-specific pretraining matters more than
general-purpose pretraining for specialized tasks.
\citet{ji2022} introduced MentalBERT and MentalRoBERTa, two masked
language models pretrained on millions of sentences from Reddit
communities centered on depression, suicide, anxiety, and related
disorders. The key finding was not just that these models
outperformed general BERT on mental health classification
benchmarks --- it was that the gap between domain-pretrained and
general-pretrained models was large enough that no amount of
fine-tuning on labeled data could close it entirely. This finding,
that the pretraining corpus shape determines what the model can
eventually learn to classify, is the direct intellectual ancestor
of our DAPT experiment. The same recipe has since been applied to
other languages: Chinese-MentalBERT followed the same approach for
Mandarin social media \citep{zhai2024}, and more recent work on
Arabic mental health text has shown that domain adaptation
generalizes across linguistic families \citep{mhg2025}. All of
this work has happened almost entirely outside Bengali, which has
been left out of the mental health NLP literature that English,
Chinese, and Arabic now have access to.

\subsection{Bengali NLP and Depression Detection}

Bengali is spoken by more than 230 million people, making it the
seventh most widely spoken language in the world. Despite this, it
remains genuinely low-resource in NLP terms, partly because of the
complexity of its morphological system and partly because
high-quality labeled datasets have been slow to accumulate
\citep{islam2021}. The foundational work for Bengali language
modeling came from two directions. \citet{bhattacharjee2022}
introduced BanglaBERT, a BERT-based model pretrained on 27.5 GB
of Bangla text crawled from 110 popular Bengali websites,
establishing the first comprehensive Bengali language understanding
benchmark (BLUB) covering text classification, sequence labeling,
and span prediction. In parallel, a collaborative effort organized
by Neuropark produced SahajBERT, an ALBERT-architecture model
pretrained on Bengali Wikipedia and the Bengali portion of OSCAR
through distributed volunteer computation \citep{sahaj2021}.
\citet{indicbert2022} introduced IndicBERTv2, a transformer
model trained on a large multilingual Indic corpus spanning 24
languages, providing broader South Asian language coverage than
MuRIL at the cost of reduced specialization for any single language
family. These models became the standard starting points for Bengali
and broader Indic language classification tasks.

The depression detection work that followed built on these
foundations. Mumu et al.\ combined CNN and LSTM layers to classify
depressive Facebook posts in Bangla \citep{mumu2023}. Uddin et al.\
compared GRU and LSTM architectures on a smaller labeled set
\citep{uddin2022}. The most complete effort, and the one our work
most directly extends, was produced by \citet{chowdhury2024}, who
built the Bengali Social Media Depressive Dataset (BSMDD) by
translating Reddit and Twitter posts into Bengali and having
native-speaking annotators with mental health expertise label them
as depressive or non-depressive. The paper tested everything from
Bi-LSTM to SahajBERT to fine-tuned GPT-3.5 and reported strong
results, with their proposed DepGPT model showing impressive
flexibility across conditions. BSMDD remains the reference point
for Bengali depression detection work, and we use it as our primary
comparison baseline throughout this paper.

The limitation that runs through most of this work, including
BSMDD, the most widely used benchmark, is that the data were built
through translation from English. The informal linguistic features that
real Bengali speakers use when expressing distress online ---
emoticons, romanized Bengali phonetic spelling, expressive
elongation --- are structurally absent from translated data,
because translators have no reason to introduce them. Code-mixed
Bengali-English text, commonly called Banglish, presents unique
challenges for standard NLP pipelines that assume monolingual
input, yet it remains underrepresented in Bengali NLP benchmarks
despite its prevalence in social media discourse \citep{roy2025}.
BanglishDepYT contains naturally occurring Banglish at a measured
rate (7.9\% of comments in the clean corpus) and releases these
informal features with the text rather than removing them, which
allows code-mixing's effect on depression detection to be studied
directly. Native collection itself is not new: \citet{ghosh2023}
collected Bengali depression text from Facebook, Twitter and
YouTube (Section~2.3). Our contribution is a native corpus whose
informal features are preserved and characterized, paired with a
preprocessing method designed to keep them.

\subsection{Preprocessing Effects in Informal Text NLP}

Preprocessing is often treated as a settled question with a
standard answer: remove stopwords, strip punctuation, delete
non-target-language tokens. For formal, edited text this approach
works reliably. For informal social media text it systematically
destroys signal. \citet{islam2020} demonstrated this concretely
for Bengali sentiment analysis, showing that preprocessing choice
produced accuracy differences large enough to matter for practical
deployment. The finding that even minor changes to tokenization
and stopword lists could shift results meaningfully suggested that
the informal Bengali text problem went deeper than choosing between
two stopword lists. \citet{roy2025} made a complementary argument
about Bengali machine translation, noting that most Bengali NLP
benchmarks were built from Wikipedia and news corpora, meaning that
informal register --- the kind found in social media --- remained
largely unaddressed and likely to produce performance surprises when
models encountered it. Neither study examined depression detection
specifically, and neither proposed a method designed to protect
emotional signal during normalization. This gap is where EPH is
positioned.

This finding aligns with a growing body of evidence across NLP
tasks suggesting that preprocessing decisions have been
systematically underestimated relative to model architecture
choices \citep{islam2020,roy2025}. Our work extends this
observation to depression detection specifically, quantifying the
preprocessing impact through the CESS framework and showing that
domain-inappropriate preprocessing can substantially lower the
performance of some model families on native informal text. Ghosh et al. \citep{ghosh2023} collected native Bengali depression text from Facebook, Twitter, and YouTube and demonstrated strong results with an attention-based hybrid architecture. Critically, their preprocessing pipeline explicitly removed emoticons and non-Bangla characters before modeling. This design choice, which was standard practice at the time, motivates our core research question: what performance is being left on the table by removing exactly the informal features that characterize genuine Bengali social media distress expression?

\subsection{Domain-Adaptive Pretraining}

The theoretical foundation for domain-adaptive pretraining was
established by \citet{gururangan2020} in ``Don't Stop
Pretraining,'' which showed across four domains and eight
classification tasks that a second phase of pretraining on
domain-specific unlabeled text consistently improved downstream
task performance, even for models that had already been pretrained
on very large general corpora. The key insight was that the gap
between a model's pretraining distribution and the target domain's
linguistic characteristics is a bottleneck that labeled data alone
cannot overcome. The more informal or specialized the target
domain, the more important domain adaptation becomes. MentalBERT
directly applied this insight to mental health NLP, using Reddit
posts as the domain corpus \citep{ji2022}. \citet{khanuja2021}
introduced MuRIL, a multilingual language model built specifically
for Indian languages including Bengali, pretrained on monolingual,
translated, and transliterated data across 17 Indian languages.
MuRIL's inclusion of transliterated text made it a natural
candidate for our experiments, since Banglish --- romanized Bengali
--- is structurally similar to transliterated Indian language text.
No prior work has applied domain-adaptive pretraining to native
Bengali social media text specifically, nor examined how it
interacts with emotion-preserving preprocessing designed for that
domain.

While domain-adaptive pretraining has consistently shown benefits
across mental health NLP tasks, the relative effectiveness of
multilingual models versus domain-adapted monolingual models on
informal low-resource text remains an open question. Our
experiments contribute empirical evidence to this debate, showing
that the relevance of pretraining data to the target domain
predicts downstream performance more reliably than the scale of
multilingual coverage alone: DAPT-SahajBERT, adapted on 163{,}217 cleaned
unlabeled Bengali social media comments, performs comparably to
MuRIL ($p=0.589$) despite starting from a much smaller monolingual
base, while IndicBERTv2, a larger multilingual model with broader
language coverage, does not significantly outperform the
unadapted monolingual baseline ($p=0.372$).

\subsection{LLM-Assisted Annotation}

Collecting labeled data for low-resource mental health tasks is
slow and expensive. Annotators need language proficiency and
sufficient familiarity with the clinical concepts being labeled to
make consistent judgments. Recent work has explored whether large
language models can substitute for or augment human annotators in
such settings. The general finding is that LLMs can produce
annotations that agree substantially with human experts on
well-defined binary classification tasks, provided the labeling
criteria are specific enough \citep{llmannotation2024}.
Cohen's kappa values in the range of 0.60 to 0.70 are commonly
reported as the boundary between moderate and substantial
human-LLM agreement, and values above this threshold are generally
taken as evidence that LLM annotations are reliable enough to
supplement gold-standard human labels in downstream training. Our
work contributes to this literature by applying LLM-assisted
annotation specifically to informal Bengali depression text using a
locally deployed model (Qwen2.5-7B via Ollama) \citep{qwen2025},
achieving a kappa of 0.611 against our dual-verified human gold
labels, and demonstrating that semi-supervised training on the
resulting pseudo-labels produces measurable improvements in AUC
($p=0.044$) over gold-label-only training. This result provides a
reproducible zero-cost annotation scaling pathway for mental health
NLP in data-scarce low-resource languages, directly applicable to
other South Asian and code-mixed language settings where expert
annotators are scarce and translation-based dataset construction
is the only current option.

\section{The BanglishDepYT Dataset}
\label{sec:dataset}

\subsection{Why Existing Datasets Were Not Enough}

Before collecting a single comment ourselves, we spent time with
what already existed. The most complete Bengali depression dataset
currently available is BSMDD, introduced by \citet{chowdhury2024},
containing 21{,}910 labeled posts split evenly between depressive
and non-depressive. On paper this looked like exactly what we
needed. When we opened the actual text, the problem became
immediately visible. Every post is written in clean, grammatically
standard Bengali. There are no emoticons. There is no Banglish.
There are no stretched-out words or repeated characters that signal
emotional intensity. This is not an accident of the subject matter.
It is a consequence of how the dataset was built. The authors took
English posts from Reddit and Twitter, had native Bengali speakers
translate them, and labeled the translations. A translator sitting
down to convert an English Reddit post into Bengali has no reason
to write ``mon kharap'' instead of \textbengali{মন খারাপ}. The
dataset looks Bengali but it does not behave like Bengali social
media text.

We also investigated whether we could patch this gap by combining
BSMDD with other available Bengali datasets that at least contain
informal language. SentNoB is a Bengali sentiment dataset built
from social media with genuinely noisy text \citep{sentnob}.
BanglaHate is a hate speech dataset from Facebook and YouTube
\citep{banglahate}. We computed a Jensen--Shannon divergence-based
dataset compatibility score (DHS) between each of these and BSMDD,
where higher values indicate more similar vocabulary
distributions. Both came
back well below what would be considered a safe threshold for
merging (0.29 and 0.33 against a threshold of 0.65). Their
vocabularies are simply too different from BSMDD's for a merged
corpus to behave consistently. Patching was not going to work. We
needed to collect native Bengali depression data directly from
where people write it.

\begin{figure}[ht]
\centering
\begin{tikzpicture}
\begin{axis}[
  xbar,
  bar width=0.40cm,
  width=0.84\columnwidth,
  height=5.8cm,
  xmin=0, xmax=60,
  symbolic y coords={
    {Negation words},
    {Elongated spelling},
    {Banglish tokens},
    {Emoticons present}
  },
  ytick=data,
  y tick label style={font=\small},
  x tick label style={font=\small},
  xlabel={Percentage (\%)},
  xlabel style={font=\small},
  legend style={
    at={(0.98,0.98)},
    anchor=north east,
    font=\small,
    legend columns=1
  },
  enlarge y limits=0.20,
  grid=major,
  grid style={dashed,gray!25},
  nodes near coords,
  nodes near coords style={font=\tiny, xshift=2pt},
  point meta=explicit symbolic,
]
\addplot[
  fill=gray!55, draw=gray!70,
  point meta=explicit symbolic
] coordinates {
  (0,{Negation words})  [0\%]
  (0,{Elongated spelling})  [0\%]
  (0,{Banglish tokens})  [0\%]
  (0,{Emoticons present})  [0\%]
};
\addplot[
  fill=ytcol!80, draw=ytcol,
  point meta=explicit symbolic
] coordinates {
  (51.7,{Negation words}) [51.7\%]
  (16.6,{Elongated spelling}) [16.6\%]
  (7.9,{Banglish tokens}) [7.9\%]
  (22.4,{Emoticons present}) [22.4\%]
};
\legend{BSMDD (translated), BanglishDepYT (native)}
\end{axis}
\end{tikzpicture}
\caption{Percentage of comments containing informal features in
BSMDD and the full BanglishDepYT clean corpus. BSMDD contains none;
these features appear naturally in comments collected directly from
YouTube. BanglishDepYT rates are computed over the full clean
corpus, not the feature-enriched annotation sample
(Section~\ref{sec:annotation}).}
\label{fig:dataset_features}
\end{figure}
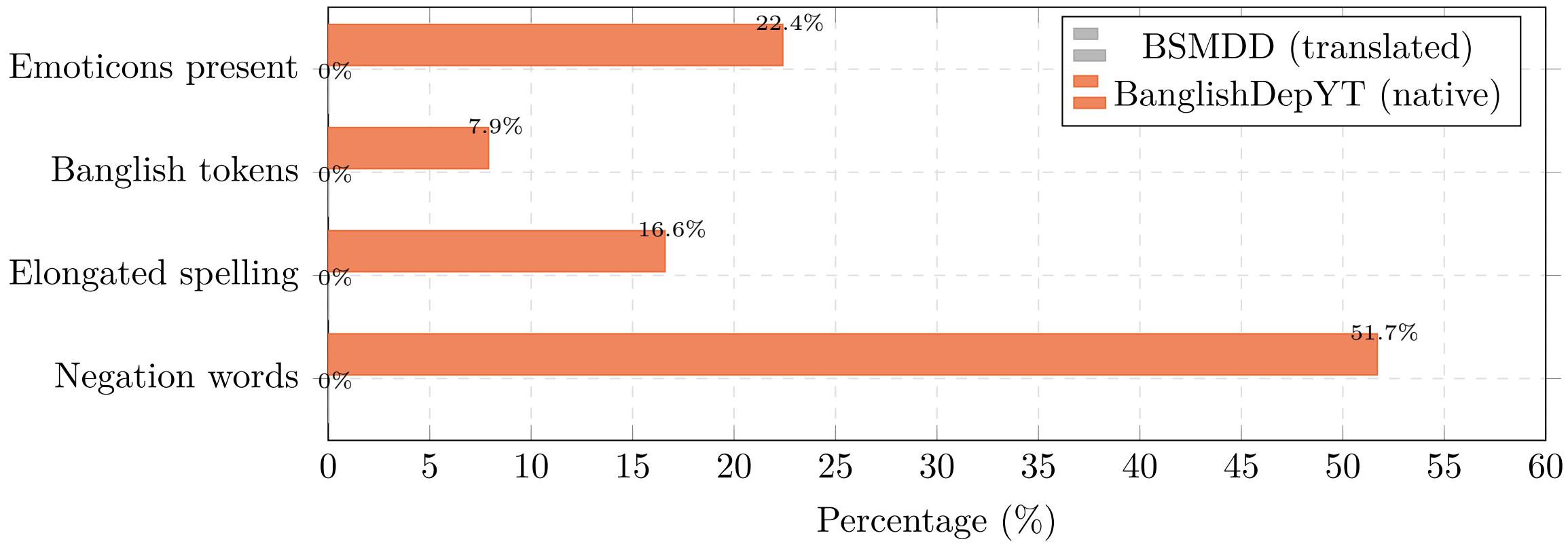

\subsection{Why YouTube}

We chose YouTube for three practical reasons. First, the Google
YouTube Data API v3 is free to use, with a daily quota large
enough to collect tens of thousands of comments without cost.
Second, Bengali mental health content on YouTube is genuinely
abundant. There are personal story channels, awareness videos,
motivational channels, reaction videos to sad songs, and comment
sections under emotional content where people write at length about
how they feel. Third, and most importantly, YouTube comments are
written the way people actually write when they are not being
formal. They use emoticons. They mix Bengali script with
phonetically spelled English. They repeat characters when they
want to express something that ordinary spelling cannot capture.
This is exactly the linguistic environment that existing Bengali
depression datasets are missing.

\subsection{Data Collection and Filtering}

We built a query bank of 150 search queries organized into 13
thematic categories. On the depressive side, categories included
direct emotional expressions in Bengali script
(\textbengali{মন খারাপ}, \textbengali{কষ্ট লাগছে},
\textbengali{আর পারছি না}), Banglish phrasing
(``mon kharap bangla,'' ``kosto lagche,'' ``ami bhalo nei''),
life situations known to correlate with depression such as
relationship breakdowns, job loss, and academic failure,
crisis-related language, physical symptoms of depression including
sleeplessness and loss of appetite, personal story confessional
content, late-night venting, and cognitive distortion language.
On the non-depressive side, we searched for happiness, motivation,
entertainment, cooking, family content, and sports, ensuring the
dataset would have meaningful contrast rather than just being a
collection of sad comments.

Running these 150 queries through the API, we identified
approximately 3{,}600 unique videos and retrieved comments from
each one. The raw collection came to 191{,}000 Bengali comments.
We applied a five-step filtering pipeline to remove noise and
low-quality content, producing a clean corpus of 163{,}217
comments that forms the foundation for every subsequent use of
this dataset. Table~\ref{tab:filtering} documents each filtering
step and its effect on corpus size. The cleaned 163{,}217-comment
corpus serves two distinct purposes in this paper: it provides
the unlabeled domain text for DAPT pretraining, and it provides
the pool from which LLM-assisted annotation was applied.

\begin{table}[ht]
\centering
\caption{Five-step filtering pipeline applied to the raw
191{,}000 YouTube comment collection. Each step addresses
a specific category of noise common in social media data.}
\label{tab:filtering}
\begin{tabularx}{\columnwidth}{lXr}
\toprule
\textbf{Step} & \textbf{Filter applied} & \textbf{Remaining} \\
\midrule
0 & Raw collection & 191{,}000 \\
1 & Remove emoji-dominant ($>$30\% emoji chars) & 188{,}741 \\
2 & Remove repetition spam ($>$6 repeated chars) & 181{,}209 \\
3 & Remove very short ($<$4 words) & 171{,}883 \\
4 & Remove very long ($>$150 words) & 165{,}102 \\
5 & Remove exact duplicates & 163{,}217 \\
\bottomrule
\multicolumn{3}{l}{\textit{Final clean corpus: 163{,}217 comments.
Used for DAPT and LLM annotation.}}\\
\end{tabularx}
\end{table}

\begin{figure}[ht]
\centering
\begin{adjustbox}{max width=\columnwidth}
\begin{tikzpicture}[
  node distance = 0.45cm and 0.9cm,
  box/.style    = {rectangle, rounded corners=4pt,
                   minimum width=3.8cm, minimum height=0.80cm,
                   align=center, font=\small,
                   draw=gray!60, thick},
  splitbox/.style = {rectangle, rounded corners=4pt,
                     minimum width=3.3cm, minimum height=0.72cm,
                     align=center, font=\scriptsize,
                     draw=gray!60, thick},
  arr/.style    = {-stealth, thick, gray!70},
  label/.style  = {font=\scriptsize, text=gray!80},
  rawcol/.style = {fill=gray!15},
  cleancol/.style= {fill=blue!10},
  goldcol/.style = {fill=orange!20},
  daptcol/.style = {fill=teal!15},
  llmcol/.style  = {fill=purple!12},
  semiscol/.style= {fill=green!12},
]

\node[box, rawcol] (raw)
  {\textbf{Raw collection}\\191{,}000 comments};

\node[box, cleancol, below=0.45cm of raw] (clean)
  {\textbf{After filtering}\\163{,}217 comments};

\draw[arr] (raw) -- (clean)
  node[label, right, midway] {5-step filter};

\node[splitbox, daptcol,
      below left=0.7cm and 3.6cm of clean] (dapt)
  {\textbf{DAPT pretraining}\\163{,}217 unlabeled\\(MLM, mask=0.15)};

\node[splitbox, goldcol, below=0.7cm of clean] (goldann)
  {\textbf{Manual annotation}\\1{,}000 comments\\(2 bilingual authors)};

\node[splitbox, llmcol,
      below right=0.7cm and 3.6cm of clean] (llm)
  {\textbf{LLM annotation}\\101{,}000 selected\\(Qwen2.5-7B)};

\draw[arr] (clean.south) -- ++(0,-0.2) -| (dapt.north);
\draw[arr] (clean) -- (goldann);
\draw[arr] (clean.south) -- ++(0,-0.2) -| (llm.north);

\node[splitbox, daptcol, below=0.45cm of dapt] (ckpt)
  {\textbf{DAPT checkpoint}\\SahajBERT adapted\\to Bengali social media};

\node[splitbox, goldcol, below=0.45cm of goldann] (goldsplit)
  {\textbf{Gold splits (EPH)}\\700 train · 150 val\\150 test};

\node[splitbox, llmcol, below=0.45cm of llm] (pseudo)
  {\textbf{Pseudo-labels}\\96{,}071 high-confidence\\66{,}851 after filtering};

\draw[arr] (dapt) -- (ckpt);
\draw[arr] (goldann) -- (goldsplit);
\draw[arr] (llm) -- (pseudo);

\node[box, daptcol, below=0.5cm of ckpt] (finetune)
  {\textbf{Fine-tuning}\\Gold 1{,}000 · 10 seeds};

\node[box, semiscol,
      below right=0.5cm and 1.8cm of goldsplit] (combined)
  {\textbf{Combined dataset}\\67{,}551 train (semi-supervised)};

\draw[arr] (ckpt) -- (finetune);
\draw[arr] (goldsplit.south) -- ++(0,-0.2) -|
  (finetune.north east);
\draw[arr] (goldsplit.south) -- ++(0,-0.2) -|
  (combined.north west);
\draw[arr] (pseudo.south) -- ++(0,-0.2) -|
  (combined.north east);

\end{tikzpicture}
\end{adjustbox}
\caption{Complete BanglishDepYT data flow. The 191{,}000 raw
comments are first filtered to 163{,}217 clean comments. This
clean corpus feeds two parallel tracks: DAPT pretraining
(all 163{,}217 unlabeled comments) and LLM-assisted annotation
(101{,}000 selected comments). Separately, 1{,}000 comments
receive manual dual-annotator labeling and form the gold
evaluation set. The pseudo-labeled and gold training sets
are combined for semi-supervised training.}
\label{fig:data_flow}
\end{figure}
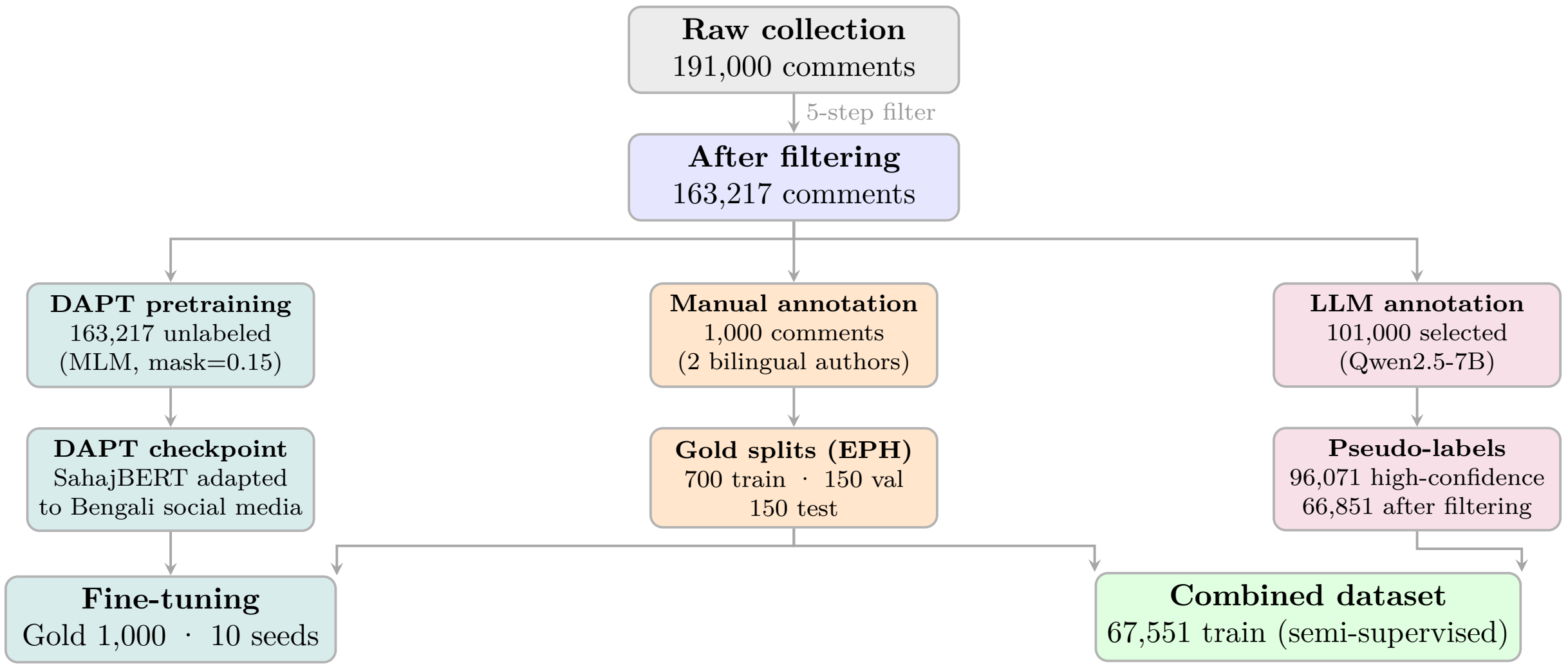

\subsection{Annotation}
\label{sec:annotation}

From the 163{,}217 clean comments, 1{,}000 were selected for
manual dual annotation using a relevance score that rewarded
the presence of unique emoticon types, Banglish word count,
elongated spelling patterns, and negation words such as
\textbengali{না}, \textbengali{নেই}, and \textbengali{নয়}. We
took the 700 highest-scoring comments and added 300 sampled at
random, so the annotation set would not consist entirely of
unusually expressive examples.

This selection procedure enriches the gold set for exactly the
features that EPH is designed to preserve. In the 1{,}000 gold
comments, 54.1\% contain emoticons, 54.4\% contain Banglish,
41.1\% contain elongation and 67.8\% contain negation words,
compared with 22.4\%, 7.9\%, 16.6\% and 51.7\% in the full clean
corpus (Figure~\ref{fig:dataset_features}). All corpus-level
feature rates reported in this paper are computed on the full
clean corpus. All classification results on the gold set,
including the EPH comparisons, are obtained on this
feature-enriched sample and may overstate the benefit of EPH on
randomly sampled comments; we discuss this further in the
Limitations.

Each of the 1{,}000 comments was read and labeled by two
bilingual annotators with native Bengali proficiency, both of
whom are authors of this paper. The annotation was not carried
out by NLP researchers alone: one of the authors, Sadman Sharif,
was trained in psychology at the Department of Educational and
Counselling Psychology, University of Dhaka, and was involved in
the annotation process, bringing a psychological perspective to
the labeling decisions. Annotation followed
a single guiding distinction: a comment is labeled depressive if
the person is expressing their own emotional state, and
non-depressive if they are commenting on something external.
``This song makes me cry every time'' was labeled non-depressive.
``I cry every night and there is nobody there'' was labeled
depressive. This distinction came up constantly because a large
share of YouTube comments under emotional videos sits exactly at
this boundary. When the two annotators disagreed, they discussed
the comment until reaching consensus. Initial disagreements were
not recorded before adjudication, so we cannot report human
inter-annotator agreement for this release. The final labeled set
contains 417 depressive comments and 583 non-depressive comments.

Because the label captures whether a commenter expresses their own
negative emotional state, not whether they meet clinical criteria
for depression, the task is more precisely described as
\emph{depressive-expression detection}. Labels were not derived
from a clinical instrument such as the PHQ-9, and a positive label
should not be read as an indication of clinical depression.
We use ``depressive'' in this narrower sense throughout.

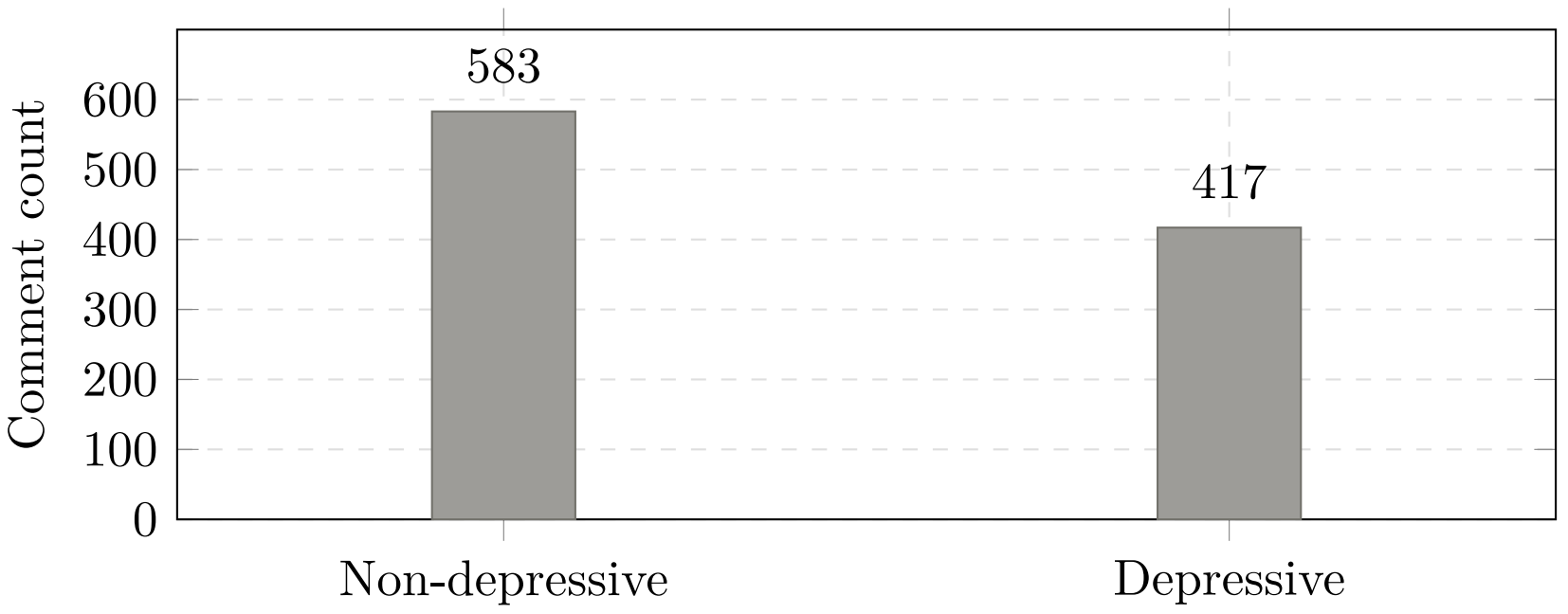
\begin{figure}[ht]
\centering
\begin{tikzpicture}
\begin{axis}[
  ybar,
  bar width=1.0cm,
  width=0.70\columnwidth,
  height=5.0cm,
  ymin=0, ymax=700,
  symbolic x coords={Non-depressive, Depressive},
  xtick=data,
  x tick label style={font=\small},
  ylabel={Comment count},
  ylabel style={font=\small},
  ytick={0,100,200,300,400,500,600},
  yticklabel style={font=\small},
  nodes near coords,
  nodes near coords style={font=\small, yshift=2pt},
  enlarge x limits=0.45,
  grid=major,
  grid style={dashed,gray!25},
]
\addplot[fill=bsmddcol!70, draw=bsmddcol]
  coordinates {(Non-depressive,583)(Depressive,417)};
\end{axis}
\end{tikzpicture}
\caption{Label distribution of the BanglishDepYT gold labeled
set (1{,}000 comments). The mild imbalance (583 non-depressive,
417 depressive) reflects the natural distribution of personal
distress expression in YouTube comment sections and is handled
through weighted loss during training rather than resampling.}
\label{fig:label_dist}
\end{figure}

\subsection{Corpus Usage: DAPT and LLM Annotation}

The 163{,}217 clean comments serve two distinct roles in this
paper and it is important to state them clearly.

\textbf{DAPT pretraining.} All 163{,}217 clean unlabeled comments
were used for domain-adaptive pretraining of SahajBERT via masked
language modeling. No labels were required for this step. The
model simply reads the comments and learns the vocabulary patterns,
Banglish spelling conventions, emoticon-adjacent language, and
informal sentence structures characteristic of Bengali mental
health social media. The scale of this corpus is what makes
domain adaptation meaningful: 163{,}217 comments provide
sufficient coverage of the target domain's linguistic variation
to shift the model's internal representations in a way that
improves downstream depression detection.

Because the 1{,}000 gold comments were drawn from this same
163{,}217-comment corpus, the DAPT corpus contains the unlabeled
text of every gold comment, including the 150 test comments.
DAPT never saw any labels, but the domain-adapted model was
exposed to the test text during masked language modeling. This is
a form of test-set contamination that may favor DAPT-SahajBERT
over models that did not see the test text, and the DAPT results
in Sections~\ref{sec:results} and~\ref{sec:stats} should be read
with this in mind. A rerun with the gold comments removed from the
DAPT corpus is required to rule out this effect.

\textbf{LLM-assisted annotation.} We applied Qwen2.5-7B
\citep{qwen2025} via Ollama to a 101{,}000-comment subset of
the clean corpus, selected after excluding the 1{,}000 manually
annotated comments and applying additional quality filters for
minimum Bengali character density. Before annotating at scale,
we validated the LLM annotator against the full 1{,}000 gold
labels. The model achieved accuracy of 0.812, weighted F1 of
0.811, and Cohen's kappa of 0.611, indicating substantial
agreement with human judgment \citep{llmannotation2024}. Of the
101{,}000 submitted comments, 96{,}071 received high-confidence
predictions: 26{,}829 depressive and 69{,}242 non-depressive.
The remaining 4{,}929 predictions (4{,}732 low-confidence and 197
unparseable) were discarded. Confidence is the model's own
\texttt{HIGH}/\texttt{LOW} self-report, requested in the prompt
and decoded at temperature 0; it is not a calibrated probability.

The 96{,}071 high-confidence pseudo-labels were then filtered
before training, in this order: (1) EPH was applied and comments
that became empty were removed; (2) any comment whose processed
text matched a gold train, validation or test comment was removed;
(3) all copies of a processed text that received conflicting
labels were removed; and (4) the remaining duplicates were
collapsed to one copy. These steps leave 66{,}851 pseudo-labeled
comments, which together with the 700 gold training comments form
the 67{,}551-comment combined training set used in our
semi-supervised experiments.

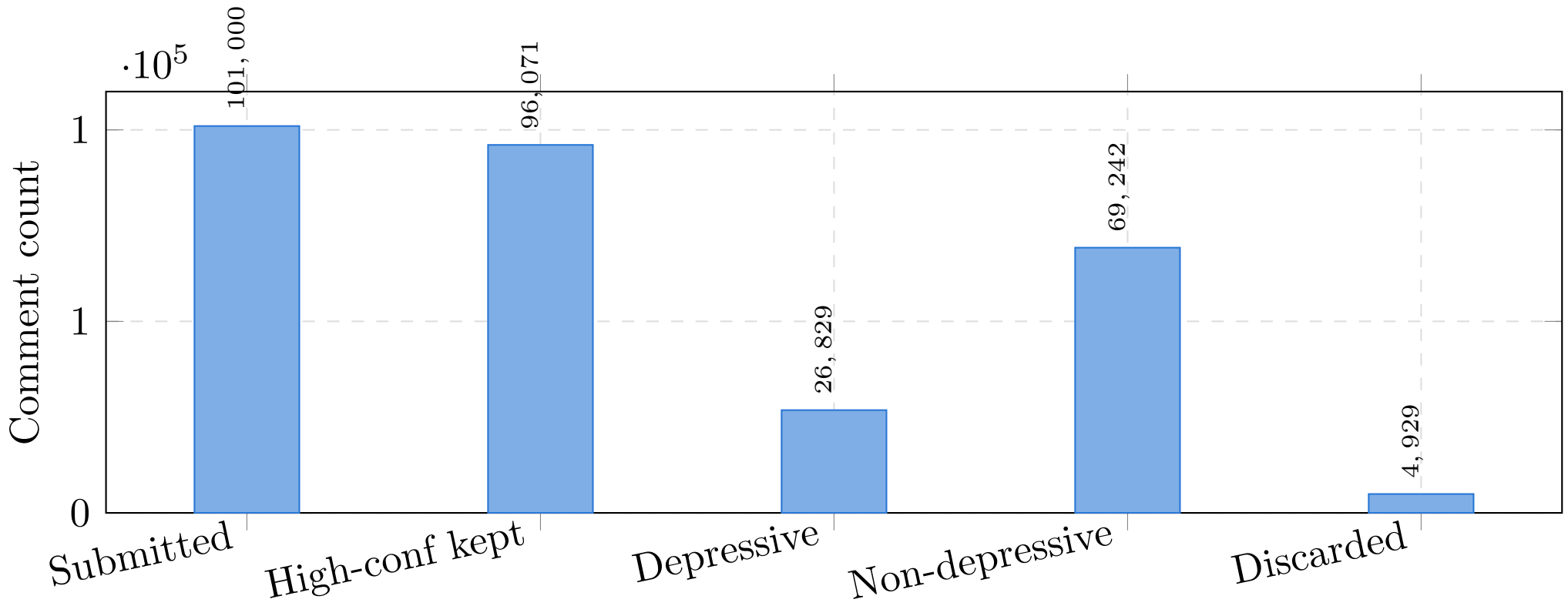
\begin{figure}[ht]
\centering
\begin{tikzpicture}
\begin{axis}[
  ybar,
  bar width=0.9cm,
  width=0.88\columnwidth,
  height=5.2cm,
  ymin=0, ymax=110000,
  symbolic x coords={
    {Submitted},
    {High-conf kept},
    {Depressive},
    {Non-depressive},
    {Discarded}
  },
  xtick=data,
  x tick label style={font=\small, rotate=12, anchor=east},
  ylabel={Comment count},
  ylabel style={font=\small},
  yticklabel style={font=\small,
    /pgf/number format/fixed,
    /pgf/number format/precision=0,
    /pgf/number format/1000 sep={,}},
  nodes near coords,
  nodes near coords style={
    font=\tiny, rotate=90,
    anchor=west, yshift=2pt,
    /pgf/number format/fixed,
    /pgf/number format/1000 sep={,}
  },
  enlarge x limits=0.12,
  grid=major,
  grid style={dashed,gray!25},
]
\addplot[fill=multicol!60, draw=multicol] coordinates {
  ({Submitted},    101000)
  ({High-conf kept}, 96071)
  ({Depressive},    26829)
  ({Non-depressive},69242)
  ({Discarded},      4929)
};
\end{axis}
\end{tikzpicture}
\caption{LLM annotation outcomes for the 101{,}000-comment
subset. Qwen2.5-7B retained 96{,}071 high-confidence predictions
(95.1\%), discarding 4{,}929 ambiguous cases. The class
distribution (26{,}829 depressive vs 69{,}242 non-depressive)
reflects the natural prevalence of personal distress expression
in YouTube mental health comment sections.}
\label{fig:llm_annotation}
\end{figure}

\subsection{LLM Annotator Validation}

Before deploying Qwen2.5-7B at scale, we validated its annotation
quality against our 1{,}000 dual-verified gold labels.
Table~\ref{tab:llm_validation} reports the full validation
metrics. A kappa of 0.611 places the LLM annotator firmly in
the range that annotation methodology literature describes as
substantial agreement, providing confidence that the resulting
pseudo-labels carry meaningful signal for downstream training
rather than random noise.

\begin{table}[ht]
\centering
\caption{LLM annotator (Qwen2.5-7B) validation against
1{,}000 human gold labels. Zero failed predictions indicates
the model produced a parseable output for every comment.}
\label{tab:llm_validation}
\begin{tabular}{lr}
\toprule
\textbf{Metric} & \textbf{Value} \\
\midrule
Accuracy          & 0.812 \\
F1 (weighted)     & 0.811 \\
Precision         & 0.811 \\
Recall            & 0.812 \\
Cohen's $\kappa$  & \textbf{0.611} \\
Failed predictions & 0 / 1{,}000 \\
\midrule
Non-dep F1        & 0.840 \\
Depressive F1     & 0.770 \\
\bottomrule
\multicolumn{2}{l}{\textit{$\kappa > 0.60$ = substantial agreement
\citep{llmannotation2024}.}}\\
\end{tabular}
\end{table}

\subsection{Dataset Statistics and What Makes It Different}

Table~\ref{tab:dataset_comparison} shows how BanglishDepYT
compares to BSMDD on the features that matter most for
depression detection in informal text.

\begin{table}[ht]
\centering
\caption{BanglishDepYT compared to BSMDD on features important
for depression detection in informal text.}
\label{tab:dataset_comparison}
\begin{tabularx}{\columnwidth}{lXX}
\toprule
\textbf{Feature} & \textbf{BSMDD}
                 & \textbf{BanglishDepYT} \\
\midrule
Total labeled    & 21{,}910  & 1{,}000 (gold) \\
Pseudo-labeled   & None      & 96{,}071 \\
Unlabeled corpus & None      & 163{,}217 \\
Source language  & English (translated) & Bengali (native) \\
Platform         & Reddit / Twitter & YouTube \\
Emoticons$^\ast$        & 0\%       & 22.4\% \\
Banglish tokens$^\ast$  & 0\%       & 7.9\%  \\
Elongated spelling$^\ast$ & Rare    & 16.6\% \\
Negation words$^\ast$   & Standard  & 51.7\% \\
Annotators       & Expert translators
                 & 2 bilingual authors \\
DAPT corpus      & None      & 163{,}217 unlabeled \\
\bottomrule
\multicolumn{3}{l}{\footnotesize $^\ast$Share of comments in the
full clean corpus; gold-set rates are higher
(Section~\ref{sec:annotation}).}\\
\end{tabularx}
\end{table}

The gold labeled set is split 70/15/15 into train, validation,
and test sets stratified by label, giving 700 training samples,
150 validation samples, and 150 test samples. The ten-seed
transformer comparison and the semi-supervised experiment in
Section~\ref{sec:results} use this split with EPH preprocessing
applied. The BanglishDepYT preprocessing experiment
(Table~\ref{tab:yt_allmodels}) uses a different split: a
class-balanced 834-comment subset of the gold set (all 417
depressive comments and 417 of the 583 non-depressive
comments), divided into 583 training, 125 validation and 126 test
comments. Results from the two splits are not directly comparable.
The combined dataset used for semi-supervised training includes
the 700 gold training samples plus the 66{,}851 filtered
pseudo-labels, a 96-fold increase in training set size over the
gold-only condition. A complete summary of all dataset subsets
and their roles is provided in Table~\ref{tab:dataset_summary}.

\begin{table}[ht]
\centering
\caption{Complete summary of all BanglishDepYT subsets and
their role in each experiment.}
\label{tab:dataset_summary}
\begin{tabularx}{\columnwidth}{lrX}
\toprule
\textbf{Subset} & \textbf{Size}
                & \textbf{Used for} \\
\midrule
Raw collection    & 191{,}000 & Starting point \\
After filtering   & 163{,}217 & DAPT + LLM annotation pool \\
DAPT corpus       & 163{,}217 & MLM pretraining of SahajBERT \\
LLM annotation    & 101{,}000 & Pseudo-label generation \\
High-conf pseudo  &  96{,}071 & Before filtering \\
Filtered pseudo   &  66{,}851 & Semi-supervised training \\
Gold labeled      &   1{,}000 & Manual annotation \\
Gold train        &     700   & Supervised fine-tuning \\
Gold val          &     150   & Validation / early stopping \\
Gold test         &     150   & Final evaluation (10-seed and
                                semi-supervised comparisons) \\
Balanced subset   &     834   & BanglishDepYT preprocessing
                                experiment (583 / 125 / 126) \\
Combined train    &  67{,}551 & Semi-supervised fine-tuning \\
\bottomrule
\end{tabularx}
\end{table}

\section{Emotion-Preserving Hybrid Preprocessing}
\label{sec:eph}

\subsection{The Problem Standard Preprocessing Creates}

Every NLP pipeline starts with a preprocessing step that decides
what to keep and what to throw away. For formal text, this
decision is straightforward. Remove the links, delete the
stopwords, strip the punctuation, and the remaining tokens carry
most of the meaning. For informal Bengali social media text, the
same logic quietly destroys exactly the signal that depression
detection depends on.

The problem is easiest to see with a concrete example. Take this
comment, which appeared in our collected corpus:

\begin{center}
\textit{``ami r parchi na \cryingemoji\cryingemoji mon kharap valo nei keu nai''}
\end{center}

A standard Bengali preprocessing pipeline does the following.
It removes \cryingemoji\cryingemoji because they are not Bengali characters. It
removes ``ami,'' ``r,'' ``parchi,'' ``na,'' ``mon,'' ``kharap,''
``valo,'' ``nei,'' ``keu,'' and ``nai'' because they either
appear on the Bengali stopword list or contain English characters.
What remains is an empty string. The comment expressed ``I cannot
take it anymore, I feel terrible, I am not well, there is nobody''
and the pipeline returned nothing. The model learns nothing from
this example.

This is not an edge case in our corpus. When we measured how much
standard preprocessing removes from BanglishDepYT's training set,
we found that the Banglish Survival Rate dropped to exactly zero.
Every single code-mixed token was deleted. On the fused
native-text corpus used for the CESS comparison
(Table~\ref{tab:cess}), the CESS score fell from 1.0 for raw text
to 0.6694 under standard preprocessing. For a depression detection task where
the most emotionally loaded words are often the Banglish ones and
the emoticons, this is not a minor calibration issue. It is a
structural failure.

EPH is built to fix this.

\subsection{The Five Preprocessing Conditions}

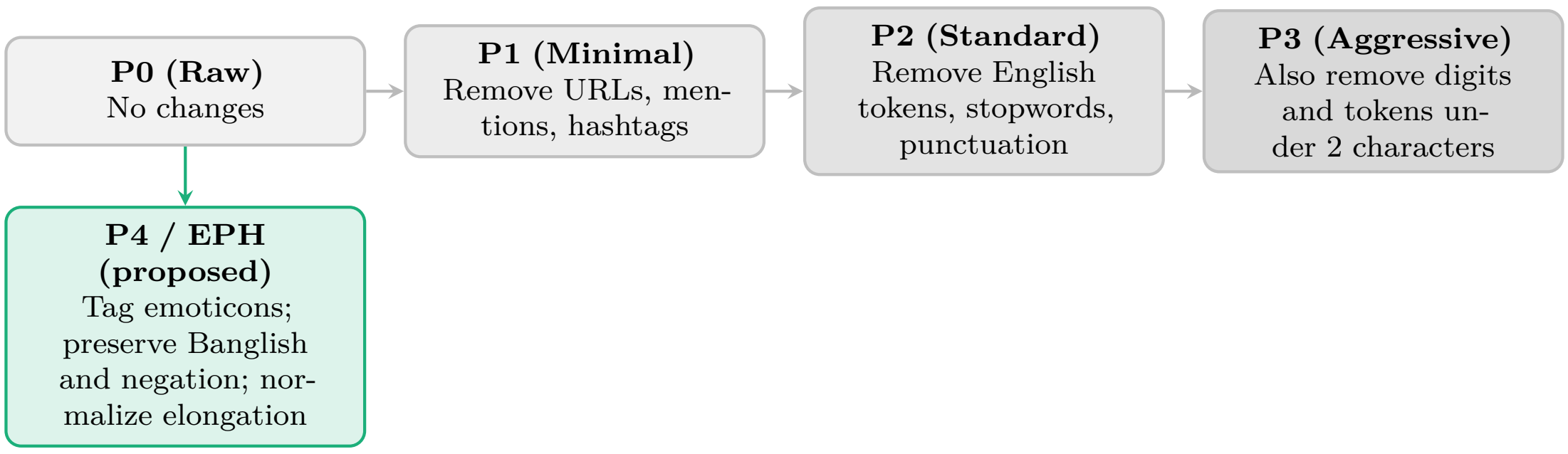
\begin{figure}[htbp]
\centering
\begin{tikzpicture}[
  node distance=0.32cm and 0.35cm,
  cond/.style={
    rectangle, rounded corners=4pt,
    text width=3.05cm, minimum height=1.0cm,
    inner sep=4pt, align=center, font=\scriptsize,
    draw=gray!50, thick
  },
  arr/.style={-stealth, thick, gray!55},
]
\node[cond, fill=gray!10] (p0)
  {\textbf{P0 (Raw)}\\No changes};
\node[cond, fill=gray!15, right=of p0] (p1)
  {\textbf{P1 (Minimal)}\\Remove URLs, mentions, hashtags};
\node[cond, fill=gray!22, right=of p1] (p2)
  {\textbf{P2 (Standard)}\\Remove English tokens, stopwords, punctuation};
\node[cond, fill=gray!30, right=of p2] (p3)
  {\textbf{P3 (Aggressive)}\\Also remove digits and tokens under 2 characters};
\node[cond, fill=daptcol!15, draw=daptcol,
      below=0.55cm of p0] (p4)
  {\textbf{P4 / EPH (proposed)}\\Tag emoticons; preserve Banglish and negation; normalize elongation};

\draw[arr] (p0.east) -- (p1.west);
\draw[arr] (p1.east) -- (p2.west);
\draw[arr] (p2.east) -- (p3.west);
\draw[arr, daptcol] (p0.south) -- (p4.north);
\end{tikzpicture}
\caption{Five preprocessing conditions as a progressive cleaning
ladder. P0 to P3 remove increasingly more content.
EPH (P4, highlighted) follows a different philosophy: it removes
noise while actively protecting emotional signal rather than
simply removing less content.}
\label{fig:preproc_ladder}
\end{figure}

Rather than simply proposing EPH and claiming it is better, we
define it within a controlled ladder of five preprocessing
conditions, each tested with four base models (SVM, BiLSTM,
BanglaBERT, SahajBERT) on BSMDD, the fused corpus and
BanglishDepYT. This design lets us isolate precisely what each
cleaning step adds or removes, and report concrete F1 differences
rather than vague claims about better or worse preprocessing.
Figure~\ref{fig:preproc_ladder} shows the five conditions as a
progressive cleaning ladder.

\textbf{P0 (Raw)} leaves the text exactly as collected.
No changes. This is the reference point for every other condition.

\textbf{P1 (Minimal)} removes only content that carries no
emotional signal in any context: URLs, user mentions beginning
with @, hashtags, and HTML tags. Everything else --- emoticons,
Banglish, elongated text, and punctuation --- is kept exactly
as written.

\textbf{P2 (Standard)} represents what most Bengali NLP work has
historically applied, and what the BSMDD preprocessing pipeline
effectively produced. It removes all non-Bengali Unicode
characters (stripping English words), deletes standard stopwords,
and strips punctuation. This is the condition EPH is most directly
competing against.

\textbf{P3 (Aggressive)} goes further than P2 by also removing
digits and discarding tokens shorter than two characters. We
include this condition to test whether cleaning more aggressively
recovers the degradation from standard preprocessing. It does not
do so: P3 scores within 0.01 F1 of P2 in all twelve model--corpus
combinations, sometimes slightly above and sometimes slightly below
(Tables~\ref{tab:bsmdd_fused_allmodels}
and~\ref{tab:yt_allmodels}).

\textbf{P4 (EPH)} is our proposed condition, described in
detail below.

\subsection{The EPH Algorithm}

EPH stands for Emotion-Preserving Hybrid preprocessing. The word
\textit{hybrid} refers to the fact that the algorithm both removes
genuinely noisy content and deliberately protects content that
carries emotional signal. Standard preprocessing only removes.
EPH removes \emph{and} protects, with protection decisions based
on how depression is expressed in informal Bengali text.

The algorithm processes text through four ordered steps.

\textbf{Step 1 --- Emoticon conversion.} Every emoticon in the
comment is converted to a plain-text semantic tag before any
cleaning begins. Table~\ref{tab:emoticon_map} lists all 17
mappings. This step matters because emoticons carry dense
emotional content in depression-related text: a string of
\cryingemoji\cryingemoji\cryingemoji is not decoration, it is distress expressed through the
fastest available shorthand. By converting emoticons to
plain-text tags, we make them visible to every downstream model.
Any remaining emoji not in our mapping are removed by an emoji
library after conversion.

\begin{table}[htbp]
\centering
\caption{Complete EPH emoticon-to-tag mapping (17 entries).
Depressive signal tags are used in CESS emotional lexicon scoring.}
\label{tab:emoticon_map}
\small
\setlength{\tabcolsep}{3pt}
\begin{tabular*}{\textwidth}{@{\extracolsep{\fill}}cllcll@{}}
\toprule
\textbf{Emoji} & \textbf{Tag} & \textbf{Signal} &
\textbf{Emoji} & \textbf{Tag} & \textbf{Signal} \\
\midrule
\cryingemoji & \texttt{[CRY]} & High-dep &
\worriedemoji & \texttt{[WORRIED]} & Medium-dep \\
\sadfaceemoji & \texttt{[SAD\_FACE]} & High-dep &
\sadfaceemoji & \texttt{[SWEAT\_SAD]} & Medium-dep \\
\pensiveemoji & \texttt{[PENSIVE]} & High-dep &
\textcolor{black}{\ensuremath{\heartsuit}} & \texttt{[BLACK\_HEART]} & Medium-dep \\
\brokenheartemoji & \texttt{[BROKEN\_HEART]} & High-dep &
\smileemoji & \texttt{[SMILE]} & Protective \\
\wearyemoji & \texttt{[WEARY]} & High-dep &
\smileemoji & \texttt{[HAPPY]} & Protective \\
\pleadingemoji & \texttt{[PLEADING]} & High-dep &
\textcolor{red}{\ensuremath{\heartsuit}} & \texttt{[HEART]} & Protective \\
\skullcrossbonesemoji & \texttt{[SKULL]} & High-dep &
\laughtearsemoji & \texttt{[LAUGH]} & Protective \\
\skullemoji & \texttt{[DEATH]} & High-dep &
\prayeremoji & \texttt{[PRAYER]} & Protective \\
\disappointedemoji & \texttt{[DISAPPOINTED]} & Medium-dep & & & \\
\bottomrule
\end{tabular*}
\end{table}

\textbf{Step 2 --- Noise removal.} URLs, user mentions,
hashtags, and HTML tags are removed. These carry no emotional
signal in any depression detection context.

\textbf{Step 3 --- Elongation normalization.} Repeated
characters are normalized to a maximum of two consecutive
repetitions. When someone types \textbengali{আআআআআ} to express
anguish, they are using repetition as an emotional amplifier.
Collapsing all repetition to a single character erases this
signal. We normalize to two rather than one because a single
character is baseline Bengali script, while a doubled character
signals emotional intensity. Three or more repetitions all
become two, standardizing the form while preserving the signal.

\textbf{Step 4 --- Token-level protection rules.} The algorithm
walks through each remaining token and applies the following
rules in priority order:

\begin{enumerate}[leftmargin=*,topsep=2pt,itemsep=2pt]
\item Semantic tags \texttt{[X]} from Step~1 --- always kept.
\item Pure punctuation or digit-only tokens --- always removed.
\item Negation words
      (\textbengali{না}, \textbengali{নেই}, \textbengali{নয়},
       \textbengali{নি}) --- always kept, because negation
      completely inverts meaning.
      \textbengali{ভালো নেই} means ``I am not well'';
      \textbengali{ভালো} alone means ``good.''
\item Emotion lexicon tokens (72 words) --- always kept.
\item Banglish tokens ($\geq$1 English char, $\geq$3 chars total)
      --- kept and lowercased.
\item Safe stopword tokens --- removed.
\item Remaining Bengali tokens ($\geq$2 chars) --- kept.
\end{enumerate}

Algorithm~\ref{alg:eph} presents the full procedure in
pseudocode.

\begin{algorithm}[htbp]
\caption{EPH Preprocessing}
\label{alg:eph}
\begin{algorithmic}[1]
\small
\Require raw comment text $t$
\Ensure preprocessed text $t'$
\For{each emoticon $e$ in \textsc{EmoticonMap}}
    \State $t \gets \text{replace}(t, e, \textsc{EmoticonMap}[e])$
\EndFor
\State $t \gets \text{remove\_remaining\_emoji}(t)$
\State $t \gets \text{remove\_noise}(t)$
      \Comment{URLs, @mentions, \#hashtags, HTML}
\State $t \gets \text{re.sub}
  (r\texttt{`(.)\textbackslash 1\{2,\}'}, \;
   r\texttt{`\textbackslash 1\textbackslash 1'}, \; t)$
      \Comment{Elongation $\to$ max 2 repeats}
\State $\textit{result} \gets [\ ]$
\For{each token $\textit{tok}$ in $\text{split}(t)$}
    \If{$\textit{tok}$ is semantic tag \texttt{[X]}}
        \State \textbf{append} $\textit{tok}$;
               \textbf{continue}
    \EndIf
    \If{$\textit{tok}$ is pure punctuation or digits}
        \State \textbf{skip}; \textbf{continue}
    \EndIf
    \If{$\textit{tok} \in \textsc{NegationWords}$}
        \State \textbf{append} $\textit{tok}$;
               \textbf{continue}
    \EndIf
    \If{$\textit{tok} \in \textsc{EmotionLexicon}$}
        \State \textbf{append} $\textit{tok}$;
               \textbf{continue}
    \EndIf
    \If{$\textit{tok}$ contains \texttt{[a-zA-Z]}
        \textbf{and} $|\textit{tok}| \geq 3$}
        \State \textbf{append} $\text{lowercase}(\textit{tok})$;
               \textbf{continue}
    \EndIf
    \If{$\textit{tok} \in \textsc{SafeStopwords}$}
        \State \textbf{skip}; \textbf{continue}
    \EndIf
    \If{$|\textit{tok}| \geq 2$}
        \State \textbf{append} $\textit{tok}$
    \EndIf
\EndFor
\State \Return $\text{join}(\textit{result})$
\end{algorithmic}
\end{algorithm}

Table~\ref{tab:eph_example} shows what happens to our example
comment under each preprocessing condition.

\begin{table}[htbp]
\centering
\caption{Preprocessing outcomes for the example comment
under each condition. Under P2 the classifier sees nothing.
Under P4 it sees negation, emotional tags, and Banglish
expressions that carry the comment's full meaning.}
\label{tab:eph_example}
\begin{tabularx}{\columnwidth}{lX}
\toprule
\textbf{Condition} & \textbf{Output} \\
\midrule
P0 (Raw)
  & \textit{ami r parchi na \cryingemoji\cryingemoji mon kharap valo nei keu nai} \\
P2 (Standard)
  & \textit{(empty --- all tokens removed)} \\
\textbf{P4 (EPH)}
  & \textbengali{পারছি না}
    \texttt{[CRY] [CRY]} \textit{mon kharap valo nei} \\
\bottomrule
\end{tabularx}
\end{table}

\subsection{Measuring Emotional Signal Preservation: The CESS
Framework}

Evaluating whether EPH preserves textual features relevant to
emotion-related classification requires an explicit measurement
framework. We introduce three diagnostic metrics and combine
them into the Composite Emotional Signal Score (CESS). These
metrics quantify the retention of selected textual features
under different preprocessing conditions. CESS is intended as
a reproducible preservation diagnostic, rather than a direct
measure of downstream semantic utility or emotional meaning.

\begin{samepage}
\textbf{Information Preservation Ratio (IPR)}\par
It measures how much entropy in the word-frequency distribution
is retained after preprocessing:
\begin{equation}
\text{IPR}(P_k) =
\min\left(1,\;\frac{H(C'_k)}{H(C)}\right)
\label{eq:ipr}
\end{equation}
\end{samepage}
where $H$ denotes Shannon entropy over word-frequency
distributions, $C$ is the original corpus, and $C'_k$ is the
corpus after preprocessing condition $P_k$. The score is capped
at 1.0. It should be interpreted as an entropy-based diagnostic,
not as a direct measure of the amount of semantic information
retained.

\textbf{Banglish Survival Rate (BSR)} measures the proportion
of identified code-mixed tokens that survive preprocessing:
\begin{equation}
\text{BSR}(P_k) =
\frac{|\{t \in B : t \text{ survives } P_k\}|}{|B|}
\label{eq:bsr}
\end{equation}
where $B$ is the set of Banglish tokens identified in the
original corpus. When $|B|=0$, as in BSMDD under our corpus
characterization, BSR is undefined. We report 1.0 by convention
to indicate that there are no identified Banglish tokens whose
loss can be measured. This convention must not be interpreted
as empirical evidence that Banglish preservation was evaluated
successfully on that corpus.

\textbf{Emotional Lexicon Preservation Score (ELPS)} measures
the retention of tokens from an emotional lexicon and the
contribution of semantic tags produced by EPH:
\begin{equation}
\text{ELPS}(P_k) =
\min\left(1,\;
\frac{|E_o \cap V'_k| + |\text{tags}(P_k)|}
     {|E_o \cap V|}
\right)
\label{eq:elps}
\end{equation}
where $E_o$ is the 72-word emotional lexicon, $V$ is the
original vocabulary, $V'_k$ is the vocabulary after preprocessing,
and $\text{tags}(P_k)$ counts the semantic tags produced by EPH.
Values are capped at 1.0. ELPS therefore reflects retention
according to the specified lexicon and tag-counting procedure;
it does not capture every possible expression of emotion.

\textbf{Composite Emotional Signal Score (CESS)} combines
the three diagnostics into a single score:
\begin{equation}
\text{CESS}(P_k) =
0.33 \cdot \text{IPR}
+ 0.33 \cdot \text{BSR}
+ 0.34 \cdot \text{ELPS}
\label{eq:cess}
\end{equation}

The weights are approximately equal and reflect the design
choice to give similar importance to entropy retention, Banglish
token survival, and emotional-lexicon preservation. These weights
are heuristic rather than empirically established contributions
to downstream classification quality. Accordingly, CESS
summarizes the selected preservation diagnostics; it does not
by itself establish their causal relationship with classifier
performance.

\subsection{CESS Results}

Table~\ref{tab:cess} reports CESS component scores for all five
preprocessing conditions on BSMDD and the fused native-text
corpus. The results describe how the defined diagnostics behave
under each condition and should be interpreted alongside the
downstream classification results.

\begin{table}[htbp]
\centering
\caption{CESS component scores on BSMDD and the fused
native-text corpus. BSR is undefined for BSMDD because no
Banglish tokens were identified and is reported as 1.0 by
convention. On the fused native-text corpus, P2 and P3 have
BSR = 0.0, whereas P4 has BSR = 1.0.}
\label{tab:cess}
\small
\resizebox{\columnwidth}{!}{%
\begin{tabular}{lrrrrrr}
\toprule
\multirow{2}{*}{\textbf{Condition}}
  & \multicolumn{3}{c}{\textbf{BSMDD}}
  & \multicolumn{3}{c}{\textbf{Fused (native)}} \\
\cmidrule(lr){2-4}\cmidrule(lr){5-7}
  & IPR & BSR$^\dagger$ & CESS
  & IPR & BSR & CESS \\
\midrule
P0 (Raw)
  & 1.0000 & 1.0000 & 1.0000
  & 1.0000 & 1.0000 & 1.0000 \\
P1 (Minimal)
  & 1.0000 & 1.0000 & 1.0000
  & 1.0000 & 1.0000 & 1.0000 \\
P2 (Standard)
  & 1.0000 & 1.0000 & 1.0000
  & 0.9983 & \textbf{0.0000} & 0.6694 \\
P3 (Aggressive)
  & 1.0000 & 1.0000 & 1.0000
  & 0.9981 & \textbf{0.0000} & 0.6694 \\
\textbf{P4 (EPH)}
  & 1.0000 & 1.0000 & 1.0000
  & 0.9987 & \textbf{1.0000} & \textbf{0.9996} \\
\midrule
\multicolumn{7}{l}{\footnotesize
$^\dagger$ BSR is undefined for BSMDD because no Banglish
tokens were identified; 1.0 is reported by convention.}\\
\multicolumn{7}{l}{\footnotesize
IPR is capped at 1.0; uncapped values above 1.0 may reflect
entropy increases associated with tag insertion.}\\
\end{tabular}%
}
\end{table}

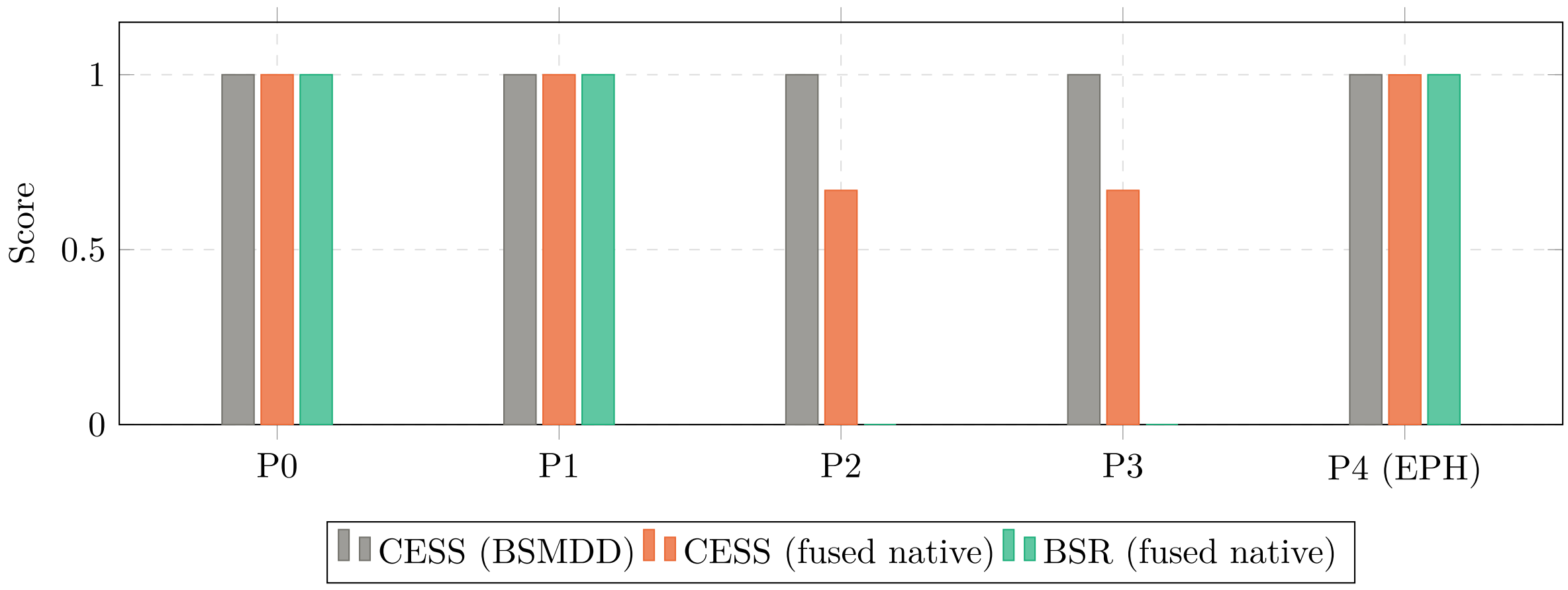
\begin{figure}[htbp]
\centering
\begin{tikzpicture}
\begin{axis}[
  ybar,
  bar width=0.32cm,
  width=\columnwidth,
  height=5.6cm,
  ymin=0.0, ymax=1.15,
  symbolic x coords={P0,P1,P2,P3,P4 (EPH)},
  xtick=data,
  x tick label style={font=\small},
  ylabel={Score},
  ylabel style={font=\small},
  yticklabel style={font=\small,
    /pgf/number format/fixed,
    /pgf/number format/precision=1},
  legend style={
    at={(0.5,-0.24)}, anchor=north,
    legend columns=3, font=\small},
  enlarge x limits=0.14,
  grid=major,
  grid style={dashed,gray!25},
]
\addplot[fill=bsmddcol!70, draw=bsmddcol]
  coordinates{
    (P0,1.0)(P1,1.0)(P2,1.0)(P3,1.0)(P4 (EPH),1.0)};
\addplot[fill=ytcol!80, draw=ytcol]
  coordinates{
    (P0,1.0)(P1,1.0)(P2,0.6694)(P3,0.6694)(P4 (EPH),0.9996)};
\addplot[fill=daptcol!70, draw=daptcol]
  coordinates{
    (P0,1.0)(P1,1.0)(P2,0.0)(P3,0.0)(P4 (EPH),1.0)};
\legend{
  CESS (BSMDD),
  CESS (fused native),
  BSR (fused native)
}
\end{axis}
\end{tikzpicture}
\caption{CESS and BSR scores across five preprocessing
conditions. BSMDD has CESS = 1.0 under the reported convention
and calculations. On the fused native-text corpus, P2 and P3
have CESS = 0.6694 and BSR = 0.0, while P4 has CESS = 0.9996
and BSR = 1.0. These values characterize the defined
preservation diagnostics and do not directly quantify the
proportion of emotional meaning retained.}
\label{fig:cess_scores}
\end{figure}

Two observations emerge from Table~\ref{tab:cess} and
Figure~\ref{fig:cess_scores}.

First, BSMDD produces CESS values of approximately 1.0 across
all five conditions, with a reported BSR of 1.0 by convention
because no Banglish tokens were identified. This pattern is
consistent with the dataset's preprocessed form and the limited
effect of the tested operations on the reported diagnostics.
However, these scores alone do not demonstrate that all five
conditions produce identical text or that previous studies
necessarily evaluated equivalent text representations. We
therefore interpret the near-constant classification scores
and preservation diagnostics as consistent with the dataset's
preprocessed form, rather than as proof of identical inputs.
This observation relates to Finding F8.

Second, on the fused native-text corpus, P2 (standard
preprocessing) and P3 (aggressive preprocessing) have BSR = 0.0,
whereas P4 (EPH) has BSR = 1.0. Under the operational definition
of BSR, this indicates that no identified Banglish tokens survive
P2 and P3, while all identified Banglish tokens survive P4.
The corresponding CESS scores are 0.6694 and 0.9996,
respectively, a difference of 0.3302.

This CESS difference is strongly influenced by the BSR component,
which has a weight of 0.33. It should not be interpreted as
evidence that standard preprocessing removes exactly one-third
of the emotional meaning retained by EPH. CESS functions as a
preservation diagnostic rather than a direct measure of
downstream semantic utility. It indicates how well preprocessing
retains the selected textual and emotional-signal proxies
defined by the metric. The downstream classification results
provide the empirical evidence for EPH's advantage, while CESS
helps characterize a potential mechanism underlying that
improvement. Accordingly, CESS should be interpreted as a
diagnostic indicator rather than proof of a causal relationship
between signal preservation and classification performance.


\section{Experimental Setup}
\label{sec:experimental_setup}

\subsection{Models}

\begin{figure}[htbp]
\centering
\resizebox{\textwidth}{!}{%
\begin{tikzpicture}[
  node distance=0.55cm and 1.4cm,
  family/.style={
    rectangle, rounded corners=6pt,
    minimum width=3.6cm, minimum height=0.75cm,
    align=center, font=\small\bfseries,
    draw=gray!60, thick
  },
  model/.style={
    rectangle, rounded corners=4pt,
    minimum width=3.4cm, minimum height=0.65cm,
    align=center, font=\small,
    draw=gray!50, thick
  },
  proposed/.style={
    rectangle, rounded corners=4pt,
    minimum width=3.4cm, minimum height=0.65cm,
    align=center, font=\small\bfseries,
    draw=daptcol, thick, fill=daptcol!12
  },
  arr/.style={-stealth, thick, gray!60},
  lbl/.style={font=\scriptsize, text=gray!70},
]

\node[family, fill=gray!12] (trad)
  {Traditional ML};
\node[model, fill=gray!8, below=0.45cm of trad] (svm)
  {SVM + TF-IDF};
\draw[arr] (trad) -- (svm);

\node[family, fill=blue!8,
      right=1.8cm of trad] (dl)
  {Deep Learning};
\node[model, fill=blue!5,
      below=0.45cm of dl] (bilstm)
  {BiLSTM + Attention\\+ FastText};
\draw[arr] (dl) -- (bilstm);

\node[family, fill=orange!10,
      right=1.8cm of dl] (mono)
  {Monolingual Transformer};
\node[model, fill=orange!7,
      below=0.45cm of mono] (bbert)
  {BanglaBERT};
\node[model, fill=orange!7,
      below=0.45cm of bbert] (sahaj)
  {SahajBERT (baseline)};
\node[proposed,
      below=0.45cm of sahaj] (dapt)
  {DAPT-SahajBERT\\(proposed)};
\draw[arr] (mono) -- (bbert);
\draw[arr] (bbert) -- (sahaj);
\draw[arr] (sahaj) -- (dapt)
  node[lbl, right, midway] {+MLM on 163k};

\node[family, fill=multicol!10,
      right=1.8cm of mono] (multi)
  {Multilingual Transformer};
\node[model, fill=multicol!7,
      below=0.45cm of multi] (muril)
  {MuRIL (17 langs)};
\node[model, fill=multicol!7,
      below=0.45cm of muril] (indic)
  {IndicBERTv2 (24 langs)};
\draw[arr] (multi) -- (muril);
\draw[arr] (muril) -- (indic);

\draw[gray!30, dashed]
  ([yshift=-0.25cm]trad.south west) --
  ([yshift=-0.25cm]multi.south east);

\end{tikzpicture}
}
\caption{Seven models evaluated in this paper organized by
architectural family. DAPT-SahajBERT (highlighted) is our
proposed model, derived from SahajBERT by continued masked
language model pretraining on 163{,}217 unlabeled Bengali
YouTube comments before supervised fine-tuning.}
\label{fig:model_families}
\end{figure}
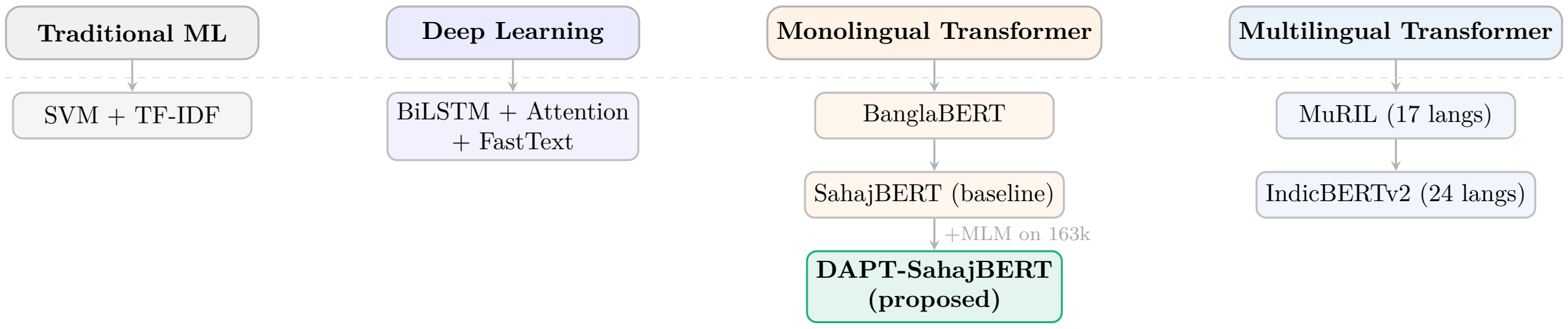

We evaluate seven models spanning three architectural families,
chosen to represent the range of approaches a practitioner would
reasonably try when building a Bengali depression detection system
from scratch. Figure~\ref{fig:model_families} shows the
architectural relationships among all seven models. Not every model
is used in every experiment. SVM, BiLSTM, BanglaBERT and SahajBERT
are the four base models of the five-condition preprocessing
comparison (Tables~\ref{tab:bsmdd_fused_allmodels}
and~\ref{tab:yt_allmodels}). SahajBERT, DAPT-SahajBERT, MuRIL and
IndicBERTv2 form the ten-seed transformer comparison on the gold
split (Table~\ref{tab:transformers}).

As a traditional machine learning baseline we use a Linear Support
Vector Machine trained on TF-IDF features with unigrams and
bigrams, a maximum of 50{,}000 features, sublinear term frequency
scaling, and a minimum document frequency of two. The SVM uses
\texttt{CalibratedClassifierCV} with three-fold cross-validation
to produce probability estimates for AUC computation. In the
BanglishDepYT preprocessing experiment, the TF-IDF vocabulary was
capped at 30{,}000 features with a minimum document frequency of
one. This model
has no notion of word order or context. It simply counts which
tokens appear and how often, making it the cleanest test of how
much information preprocessing removes or preserves, since the SVM
cannot compensate for missing tokens the way contextual models can.

As a sequence-based deep learning model we use a Bidirectional
LSTM with an attention pooling layer fed by pretrained Bengali
FastText embeddings \citep{fasttext2017}. The FastText model was
pretrained on Bengali Common Crawl data with 300-dimensional
embeddings. Unlike the SVM, this model reads text left to right
and right to left simultaneously, building a representation of
word order and short-range dependencies before making a
prediction. The attention layer weights each token's contribution
to the final representation, allowing emotionally loaded tokens
to receive more influence over the classification decision than
neutral filler tokens.

For transformer-based models we test four pretrained encoders.
BanglaBERT \citep{bhattacharjee2022} is a monolingual Bengali
transformer pretrained on 27.5 GB of crawled Bangla text using
a discriminator-based objective. SahajBERT \citep{sahaj2021}
is a monolingual Bengali ALBERT-architecture model trained
through a distributed volunteer computation experiment on Bengali
Wikipedia and the OSCAR Bengali corpus. MuRIL \citep{khanuja2021}
is a multilingual transformer from Google trained specifically on
17 Indian languages including Bengali, with dedicated training on
transliterated text that makes it a natural candidate for
Banglish-containing data. IndicBERTv2 \citep{indicbert2022} is a
more recent Indic-language transformer trained on a larger and
more diverse South Asian language corpus, providing a second
multilingual comparison point.

Finally, we introduce DAPT-SahajBERT, our proposed model. Rather
than fine-tuning SahajBERT directly on labeled examples, we first
continue its pretraining on all 163{,}217 unlabeled clean YouTube
comments using masked language modeling. This domain-adaptive
pretraining step allows the model to absorb the vocabulary,
Banglish spelling patterns, emoticon-adjacent language, and
informal sentence structures of Bengali mental health social media
before it ever sees a depression label. We then fine-tune the
resulting checkpoint on the same labeled data used for every other
transformer under identical hyperparameters, so that performance
differences reflect the domain pretraining step. Because the
pretraining corpus contains the unlabeled text of the gold test
comments, part of any gain could come from exposure to the test
text itself (Section~\ref{sec:dataset}).

\subsection{Datasets and Experimental Conditions}

We run experiments across three corpora and five preprocessing
conditions, though not every model runs on every combination.
Table~\ref{tab:experimental_conditions} summarizes which
experiments run where and why.

The full preprocessing comparison runs five conditions across all
four base models on BSMDD and on our fused corpus combining BSMDD
with SentNoB and BanglaHate using DHS-proportional weights. These
two corpora are large enough to support stable five-condition
comparisons and provide the BSMDD versus native corpus contrast
that motivates the paper's central argument. For the fused corpus,
SentNoB negative-sentiment labels and BanglaHate hate-speech labels
were mapped to the depressive proxy class, with positive, neutral,
and non-hate labels mapped to the non-depressive proxy; this
mapping is a heuristic approximation used solely for the
preprocessing signal-preservation comparison and not for any
depression classification claim.\footnote{The fused corpus is
used only to evaluate CESS and SVM preprocessing effects on
text containing native informal features. All depression
classification results in Section~\ref{sec:results} use only
the BanglishDepYT gold set with verified depression labels.}
The same five-condition comparison with all four base models also
runs on BanglishDepYT directly, using the class-balanced
834-comment subset of the gold set (583 train, 125 validation,
126 test) and three seeds. This experiment produces the
fifteen-point SVM gain from EPH; its test split differs from the
700/150/150 gold split used for the ten-seed transformer
comparison.

The DAPT comparison and multi-model transformer evaluation run
exclusively on the BanglishDepYT gold dataset. This is the only
corpus where domain-adaptive pretraining on matched unlabeled text
is meaningful, since the 163{,}217 clean unlabeled comments come
from the same YouTube distribution as the labeled gold set.

The semi-supervised training comparison runs on BanglishDepYT
splits, since the pseudo-labeled data was annotated from the same
YouTube corpus.

\begin{table}[htbp]
\centering
\caption{Experimental conditions and datasets used for each
principal comparison.}
\label{tab:experimental_conditions}
\small
\begin{tabularx}{\columnwidth}{
  >{\raggedright\arraybackslash}p{2.4cm}
  >{\raggedright\arraybackslash}p{8.5cm}
  >{\raggedright\arraybackslash}X}
\toprule
\textbf{Experiment}
  & \textbf{Datasets and conditions}
  & \textbf{Purpose} \\
\midrule
Preprocessing comparison
  & BSMDD and DHS-fused corpus;
    five conditions; four base models
  & Preprocessing effects on translated vs native text \\
Native preprocessing comparison
  & BanglishDepYT balanced subset (583/125/126); five conditions;
    four base models; 3 seeds
  & EPH gain on native YouTube data \\
DAPT and transformer
  & BanglishDepYT gold (700/150/150); SahajBERT, DAPT-SahajBERT,
    MuRIL, IndicBERTv2; 10 seeds
  & Model comparison on matched domain \\
Semi-supervised
  & BanglishDepYT; gold-only vs
    gold + pseudo-labels; 5 seeds
  & Effect of LLM pseudo-label scaling \\
\bottomrule
\end{tabularx}
\end{table}
\subsection{Training Protocol}

All models use a 70/15/15 train, validation, test split
stratified by label. Gold dataset splits are fixed at 700
training, 150 validation, and 150 test samples; the BanglishDepYT
preprocessing comparison instead uses the 583/125/126 split of the
balanced 834-comment subset described in
Section~\ref{sec:dataset}. The combined
dataset for semi-supervised training uses the same 150 test
examples as the gold condition, so all test evaluations are
directly comparable.

For the preprocessing comparison we train each model three times
with seeds 42, 123, and 456. For the DAPT and multi-model
transformer comparison we use ten seeds: 42, 123, 999, 2024, 7,
88, 314, 777, 1001, and 5555. For the semi-supervised comparison
we use five seeds, balancing statistical reliability against the
longer training time of the 67{,}551-sample combined set.

All transformer models are fine-tuned under identical
hyperparameters. We use AdamW with learning rate $2\times10^{-5}$,
batch size 32 (train) and 64 (eval), up to 8 epochs, early
stopping after 3 consecutive epochs of no improvement on
validation F1, linear warmup over 10\% of training steps, weight
decay 0.01, and mixed-precision (fp16) training. Using identical
hyperparameters for all four transformers is an intentional design
choice: we are comparing what the models are, not how well each
can be individually tuned.

The BiLSTM uses Adam with learning rate $10^{-3}$, cosine
annealing scheduling, class-weighted cross-entropy loss, and early
stopping after five epochs. The SVM uses $C=1.0$ with three-fold
internal cross-validation for probability calibration. LinearSVC
is deterministic given a fixed random state, explaining why SVM
results show zero standard deviation throughout our tables.

DAPT pretraining continued SahajBERT's masked language modeling
on all 163{,}217 clean unlabeled YouTube comments preprocessed
with EPH, with masking probability 0.15, using the AdamW
optimizer with learning rate $5\times10^{-5}$, batch size 32,
maximum sequence length 128, and linear warmup over the first
10\% of steps. Training ran for 3 epochs (approximately
15{,}300 steps) and took 4.5 hours on the RTX 4080 Super.
The domain-pretrained checkpoint was saved and used to initialize
DAPT-SahajBERT fine-tuning. The pretraining and fine-tuning phases
are separated in the sense that no labels are used during
pretraining, but the pretraining corpus includes the unlabeled
text of the gold validation and test comments
(Section~\ref{sec:dataset}).

\begin{figure}[htbp]
\centering
\begin{tikzpicture}[
  node distance=0.50cm and 1.4cm,
  stage/.style={
    rectangle, rounded corners=5pt,
    minimum width=3.8cm, minimum height=0.72cm,
    align=center, font=\small,
    draw=gray!55, thick
  },
  arr/.style={-stealth, thick, gray!60},
  lbl/.style={font=\scriptsize, text=gray!65},
]

\node[stage, fill=teal!12] (unlabeled)
  {163{,}217 unlabeled comments\\(EPH preprocessed)};

\node[stage, fill=teal!20, below=0.50cm of unlabeled] (mlm)
  {Masked language modeling\\(mask prob.\ $= 0.15$)};

\node[stage, fill=teal!30, below=0.50cm of mlm] (ckpt)
  {DAPT-SahajBERT checkpoint\\(domain-adapted)};

\draw[arr] (unlabeled) -- (mlm)
  node[lbl, right, midway] {SahajBERT base};
\draw[arr] (mlm) -- (ckpt);

\node[stage, fill=orange!12,
      right=1.8cm of unlabeled] (gold)
  {Gold labeled set\\700 train · 150 val · 150 test};

\node[stage, fill=orange!20,
      below=0.50cm of gold] (split)
  {70 / 15 / 15 stratified split\\(EPH preprocessed)};

\draw[arr] (gold) -- (split);

\node[stage, fill=daptcol!18,
      below=1.10cm of ckpt] (ft)
  {Fine-tuning\\AdamW · LR $2\!\times\!10^{-5}$ · 8 epochs\\
   Early stopping (patience 3)};

\draw[arr] (ckpt.south) -- ++(0,-0.35) -| (ft.north west);
\draw[arr] (split.south) -- ++(0,-0.35) -| (ft.north east);

\node[stage, fill=gray!12,
      below=0.50cm of ft] (eval)
  {Evaluation on 150 test samples\\
   F1 · MCC · AUC · Paired $t$-test · McNemar};

\draw[arr] (ft) -- (eval);

\draw[arr, dashed, gray!45]
  ([xshift=0.3cm]split.south) -- ++(0,-0.55)
  node[lbl, right] {Baseline: SahajBERT init.};

\end{tikzpicture}
\caption{DAPT-SahajBERT training pipeline. The left branch
adapts SahajBERT to the target domain through masked language
modeling on 163{,}217 unlabeled comments before fine-tuning.
The right branch provides the labeled splits. Both feed into
fine-tuning under identical hyperparameters. The dashed line
indicates the baseline condition, which skips DAPT and
fine-tunes SahajBERT directly from its original checkpoint.}
\label{fig:training_pipeline}
\end{figure}
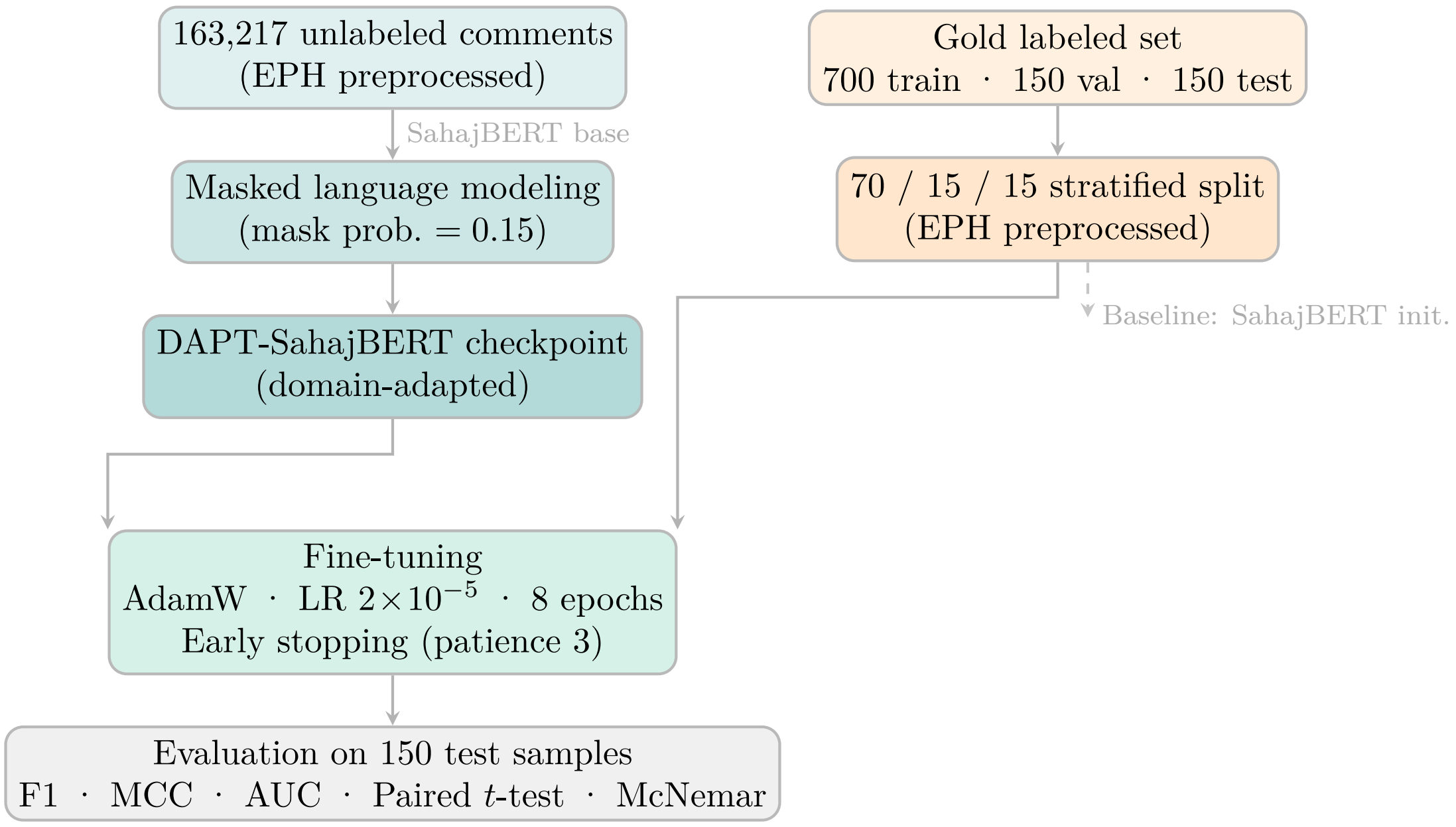

\subsection{Evaluation Metrics}

We report accuracy, weighted precision, weighted recall, and
weighted F1 for all experiments, using weighted averaging to
account for the mild label imbalance in BanglishDepYT. We
additionally report the Matthews Correlation Coefficient (MCC)
throughout \citep{matthews1975}, which accounts for all four
cells of the confusion matrix and penalizes models that achieve
high F1 by predicting the majority class more than they should.
For depression screening, where missing a true depressive case
carries real costs, MCC provides a more complete picture of
classification quality than F1 alone. We report the area under
the ROC curve (AUC) as a threshold-independent measure of
ranking quality, which matters in deployment contexts where
the decision threshold may be adjusted based on sensitivity
requirements.

For our central DAPT versus baseline comparison and the
cross-model comparisons, we go beyond mean scores. We run paired
$t$-tests across the ten per-seed F1 scores and report the paired
effect size $d_z$ (mean per-seed difference divided by its
standard deviation; \citealp{cohen1988}). Because every seed is
evaluated on the same 150 test comments, these tests measure
whether a difference is robust to training randomness
(initialization and data order); they do not test whether it
generalizes to new data. We therefore also run McNemar's test on
paired test-set predictions at seed 42 \citep{mcnemar1947}, which
tests at the level of individual test comments whether DAPT
corrects more errors than it introduces. For the semi-supervised
comparison we run paired $t$-tests across five per-seed scores
for F1, MCC, and AUC.


\section{Results}
\label{sec:results}

\subsection{Does Preprocessing Choice Actually Matter?}

\begin{figure}[htbp]
\centering
\begin{tikzpicture}
\begin{axis}[
  ybar,
  bar width=0.32cm,
  width=\columnwidth,
  height=5.8cm,
  ymin=0.60, ymax=0.92,
  symbolic x coords={P0 (Raw), P1 (Min), P2 (Std),
                     P3 (Agg), P4 (EPH)},
  xtick=data,
  x tick label style={font=\small},
  ylabel={F1 score},
  ylabel style={font=\small},
  yticklabel style={font=\small,
    /pgf/number format/fixed,
    /pgf/number format/precision=2},
  legend style={
    at={(0.5,-0.22)}, anchor=north,
    legend columns=2, font=\small},
  enlarge x limits=0.12,
  grid=major,
  grid style={dashed,gray!25},
  nodes near coords,
  nodes near coords style={
    font=\tiny, rotate=90,
    anchor=west, yshift=2pt},
  point meta=explicit symbolic,
]
\addplot[fill=bsmddcol!80, draw=bsmddcol,
         point meta=explicit symbolic]
  coordinates{
    (P0 (Raw), 0.8408)[0.8408]
    (P1 (Min), 0.8408)[0.8408]
    (P2 (Std), 0.8407)[0.8407]
    (P3 (Agg), 0.8401)[0.8401]
    (P4 (EPH), 0.8404)[0.8404]};
\addplot[fill=ytcol!80, draw=ytcol,
         point meta=explicit symbolic]
  coordinates{
    (P0 (Raw), 0.7458)[0.7458]
    (P1 (Min), 0.7301)[0.7301]
    (P2 (Std), 0.7062)[0.7062]
    (P3 (Agg), 0.7063)[0.7063]
    (P4 (EPH), 0.8570)[0.8570]};
\legend{BSMDD (translated), YouTube (native)}
\end{axis}
\end{tikzpicture}
\caption{SVM F1 across five preprocessing conditions on BSMDD
and BanglishDepYT (YouTube). BSMDD scores vary by only 0.0007,
whereas BanglishDepYT achieves F1 scores of 0.7458 with raw
text (P0), 0.7062 with standard preprocessing (P2), and 0.8570
with EPH (P4). The EPH improvements are 0.1112 over raw text
and 0.1508 over standard preprocessing in this experiment.}
\label{fig:preprocessing}
\end{figure}
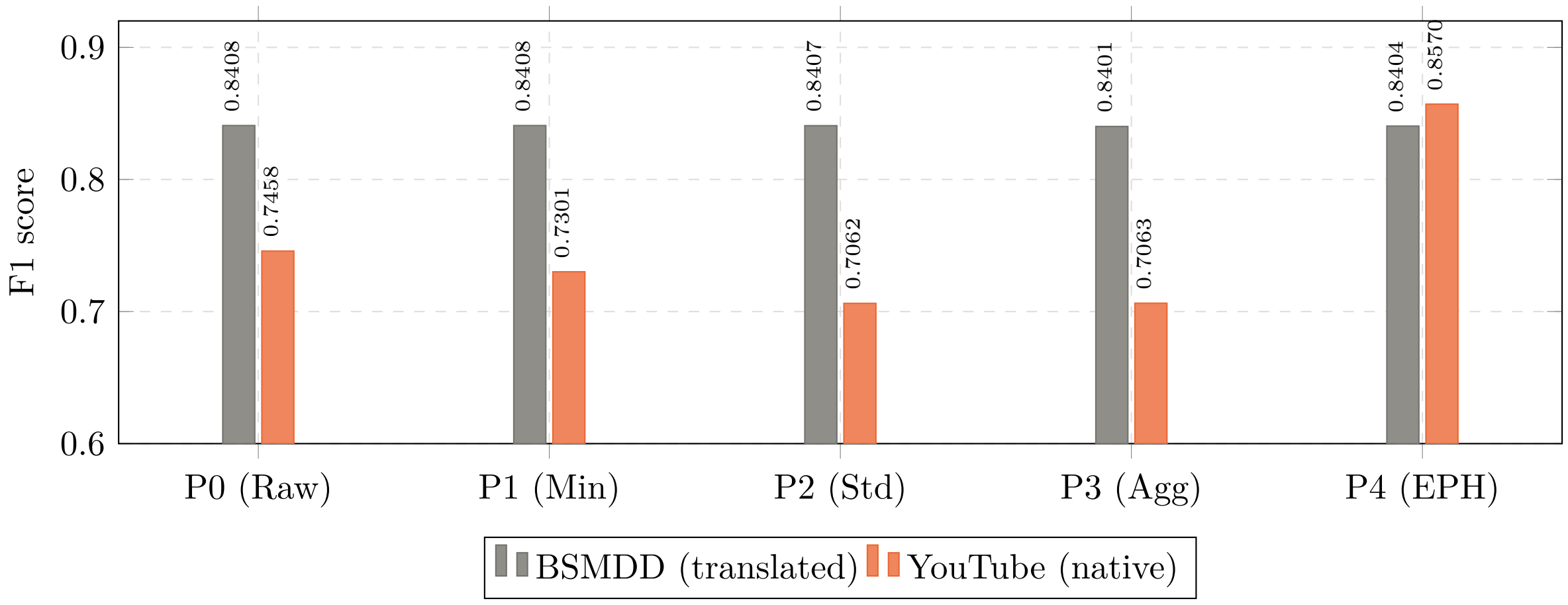

The results suggest that the effect of preprocessing depends on
the characteristics of the corpus. Table~\ref{tab:preprocessing}
and Figure~\ref{fig:preprocessing} report SVM F1 scores across
five preprocessing conditions on two different corpora. On
BSMDD, the scores remain nearly constant, ranging from 0.8401
to 0.8408. In contrast, on our native YouTube corpus, performance
varies substantially across preprocessing conditions. EPH
achieves an F1 score of 0.8570, compared with 0.7458 for raw
text (P0) and 0.7062 for standard preprocessing (P2).

\begin{table}[htbp]
\centering
\caption{SVM F1 scores across five preprocessing conditions on
BSMDD and BanglishDepYT (YouTube). Difference is the absolute
cross-corpus F1 gap for each condition. YouTube scores are on the
126-comment test split of the balanced 834-comment subset, not
the 150-comment gold test split of Table~\ref{tab:transformers}.}
\label{tab:preprocessing}
\setlength{\tabcolsep}{4pt}
\begin{tabular}{lrrr}
\toprule
\textbf{Condition} & \textbf{BSMDD}
                   & \textbf{YouTube} & \textbf{Difference} \\
\midrule
P0 (Raw)          & \textbf{0.8408} & 0.7458 & 0.0950 \\
P1 (Minimal)      & \textbf{0.8408} & 0.7301 & 0.1107 \\
P2 (Standard)     & 0.8407          & 0.7062 & 0.1345 \\
P3 (Aggressive)   & 0.8401          & 0.7063 & 0.1338 \\
\textbf{P4 (EPH)} & 0.8404          & \textbf{0.8570} & \textbf{0.0166} \\
\bottomrule
\end{tabular}
\par\smallskip
\begin{minipage}{\linewidth}
\footnotesize\raggedright
Difference = BSMDD $-$ YouTube. Bold identifies the highest value
in each score column and the smallest cross-corpus difference.
\end{minipage}
\end{table}

The near-constant BSMDD scores are consistent with the corpus
having been released in preprocessed form. Its F1 scores vary by
only 0.0007 across the five conditions. This stability suggests
that the tested preprocessing operations have little effect on
SVM performance on this corpus. However, the scores alone do
not establish that all five conditions produce identical input
text or that previous studies necessarily tested equivalent
representations. These explanations would require direct
comparison of the resulting text.

The YouTube results show a different pattern. EPH improves
F1 by 0.1112 over raw text (P0) and by 0.1508 over standard
preprocessing (P2). These comparisons address different
baselines: P0 measures the change relative to leaving the text
unprocessed, whereas P2 provides the more direct comparison
with conventional cleaning. The results are consistent with the
motivation behind EPH: preserving code-mixed expressions,
protecting negation, converting emoticons into semantic tags,
and normalizing emotional elongation may help retain information
useful for depression-related classification. The F1 scores
support the effectiveness of the complete EPH pipeline in this
SVM experiment, but they do not isolate the contribution of
each individual component. Section~\ref{sec:allmodels} shows that
the size of the EPH effect depends strongly on the model.

The cross-corpus comparison also requires care. Under P0--P3,
BSMDD achieves higher F1 than the YouTube corpus, with gaps
ranging from 0.0950 to 0.1345. Under P4, the YouTube score
exceeds the BSMDD score, and the absolute gap narrows to
0.0166. This indicates that the observed performance difference
between the corpora depends on the preprocessing condition.
Because the corpora differ in their source and text
characteristics, these results do not establish that preprocessing
is generally more important than model selection. Rather, they
show that preprocessing choice can substantially affect
performance for the evaluated SVM model on native, informal
YouTube text.

For the SVM on YouTube text, P3 (aggressive cleaning) performs
at the same low level as P2 (0.7063 versus 0.7062). Across the
other models and corpora, P3 stays within 0.01 F1 of P2
(Section~\ref{sec:allmodels}), so aggressive cleaning neither
recovers nor substantially worsens the loss from standard
preprocessing.

\subsection{Preprocessing Effects Across Model Families}
\label{sec:allmodels}

Tables~\ref{tab:bsmdd_fused_allmodels} and~\ref{tab:yt_allmodels}
report the full five-condition comparison for all four base models
on BSMDD, the fused corpus, and BanglishDepYT (three seeds each;
SVM is deterministic and therefore has zero standard deviation).

\begin{table}[htbp]
\centering
\caption{Weighted F1 (mean\,$\pm$\,std over seeds 42, 123, 456)
for four base models under five preprocessing conditions on BSMDD
and on the fused corpus (BSMDD + SentNoB + BanglaHate). Bold marks
the best condition per model and corpus.}
\label{tab:bsmdd_fused_allmodels}
\scriptsize
\setlength{\tabcolsep}{3pt}
\resizebox{\textwidth}{!}{%
\begin{tabular}{llccccc}
\toprule
\textbf{Corpus} & \textbf{Model} & \textbf{P0} & \textbf{P1}
  & \textbf{P2} & \textbf{P3} & \textbf{P4 (EPH)} \\
\midrule
\multirow{4}{*}{BSMDD}
 & SVM        & \textbf{0.8408\,$\pm$\,0.0000} & \textbf{0.8408\,$\pm$\,0.0000} & 0.8407\,$\pm$\,0.0000 & 0.8401\,$\pm$\,0.0000 & 0.8404\,$\pm$\,0.0000 \\
 & BiLSTM     & \textbf{0.8778\,$\pm$\,0.0044} & \textbf{0.8778\,$\pm$\,0.0044} & 0.8768\,$\pm$\,0.0017 & 0.8745\,$\pm$\,0.0063 & 0.8749\,$\pm$\,0.0035 \\
 & BanglaBERT & \textbf{0.8854\,$\pm$\,0.0005} & \textbf{0.8854\,$\pm$\,0.0005} & 0.8772\,$\pm$\,0.0042 & 0.8781\,$\pm$\,0.0009 & 0.8795\,$\pm$\,0.0031 \\
 & SahajBERT  & 0.9043\,$\pm$\,0.0026 & 0.9043\,$\pm$\,0.0026 & 0.9057\,$\pm$\,0.0031 & \textbf{0.9099\,$\pm$\,0.0036} & 0.9063\,$\pm$\,0.0033 \\
\midrule
\multirow{4}{*}{Fused}
 & SVM        & \textbf{0.7992\,$\pm$\,0.0000} & 0.7987\,$\pm$\,0.0000 & 0.7943\,$\pm$\,0.0000 & 0.7941\,$\pm$\,0.0000 & 0.7956\,$\pm$\,0.0000 \\
 & BiLSTM     & \textbf{0.8451\,$\pm$\,0.0024} & 0.8426\,$\pm$\,0.0066 & 0.8393\,$\pm$\,0.0030 & 0.8409\,$\pm$\,0.0048 & 0.8448\,$\pm$\,0.0013 \\
 & BanglaBERT & \textbf{0.8565\,$\pm$\,0.0038} & 0.8547\,$\pm$\,0.0055 & 0.8490\,$\pm$\,0.0058 & 0.8497\,$\pm$\,0.0030 & 0.8550\,$\pm$\,0.0009 \\
 & SahajBERT  & 0.8764\,$\pm$\,0.0034 & \textbf{0.8776\,$\pm$\,0.0023} & 0.8693\,$\pm$\,0.0010 & 0.8724\,$\pm$\,0.0010 & 0.8733\,$\pm$\,0.0031 \\
\bottomrule
\end{tabular}%
}
\end{table}

\begin{table}[htbp]
\centering
\caption{Weighted F1 (mean\,$\pm$\,std over seeds 42, 123, 456)
for four base models under five preprocessing conditions on
BanglishDepYT (balanced 834-comment subset, 126-comment test
split). Bold marks the best condition per model.}
\label{tab:yt_allmodels}
\scriptsize
\setlength{\tabcolsep}{3pt}
\resizebox{\textwidth}{!}{%
\begin{tabular}{lccccc}
\toprule
\textbf{Model} & \textbf{P0} & \textbf{P1}
  & \textbf{P2} & \textbf{P3} & \textbf{P4 (EPH)} \\
\midrule
SVM        & 0.7458\,$\pm$\,0.0000 & 0.7301\,$\pm$\,0.0000 & 0.7062\,$\pm$\,0.0000 & 0.7063\,$\pm$\,0.0000 & \textbf{0.8570\,$\pm$\,0.0000} \\
BiLSTM$^\ddagger$ & 0.6448\,$\pm$\,0.2708 & 0.6469\,$\pm$\,0.2724 & \textbf{0.6737\,$\pm$\,0.2948} & 0.6655\,$\pm$\,0.2878 & 0.6374\,$\pm$\,0.2636 \\
BanglaBERT & 0.8543\,$\pm$\,0.0046 & \textbf{0.8622\,$\pm$\,0.0048} & 0.8355\,$\pm$\,0.0166 & 0.8411\,$\pm$\,0.0158 & 0.8621\,$\pm$\,0.0124 \\
SahajBERT  & 0.8330\,$\pm$\,0.0137 & 0.8251\,$\pm$\,0.0277 & \textbf{0.8411\,$\pm$\,0.0081} & 0.8343\,$\pm$\,0.0091 & 0.8409\,$\pm$\,0.0079 \\
\bottomrule
\end{tabular}%
}
\par\smallskip
\begin{minipage}{\linewidth}
\footnotesize\raggedright
$^\ddagger$At seed 42 the BiLSTM collapsed to majority-class
prediction (F1\,=\,0.3333) under every condition; its means are
therefore unreliable.
\end{minipage}
\end{table}

Three patterns emerge. First, on BSMDD every model varies by less
than 0.01 F1 across conditions, and P0 and P1 give identical scores
for all four models. This is consistent with the observation in
Section~\ref{sec:eph} that BSMDD was released in preprocessed form.

Second, on the fused corpus, EPH scores slightly above standard
preprocessing for all four models (P4$-$P2 between $+0.0013$ and
$+0.0060$), but no model does best under EPH: the best condition
is raw or minimally processed text (P0 or P1) in every case.

Third, on BanglishDepYT the effect of EPH depends strongly on the
model. Relative to standard preprocessing, EPH improves the SVM by
0.1508 and BanglaBERT by 0.0266, leaves SahajBERT unchanged
($-0.0002$), and lowers the BiLSTM by 0.0364, although the BiLSTM
figures are unreliable because of training collapse. Relative to
raw text, the gains are 0.1112 for the SVM and under 0.01 for both
transformers. On this split the SVM with EPH (0.8570) does not
outperform the transformers: BanglaBERT reaches 0.8621 with EPH and
0.8622 with minimal preprocessing. The large SVM gain therefore
mainly shows that a bag-of-words model depends heavily on which
tokens survive preprocessing, while pretrained transformers are
much less sensitive to it. One reason transformers may benefit less
is that EPH's semantic tags such as \texttt{[CRY]} were not added
to the transformer vocabularies as special tokens, so subword
tokenizers split each tag into several pieces. Adding the tags as
special tokens is a direct next step.

\textbf{Dataset fusion gain.} We define the dataset fusion gain as
$\text{DFG} = \text{F1}_{\text{fused}} - \text{F1}_{\text{BSMDD}}$
for each model and condition. DFG is negative in all 20
model--condition pairs of Table~\ref{tab:bsmdd_fused_allmodels},
ranging from $-0.0246$ to $-0.0464$: adding SentNoB and BanglaHate
with heuristic proxy labels lowered performance for every model
and preprocessing condition. Both auxiliary datasets had a DHS
below the 0.65 threshold (0.29 and 0.33;
Section~\ref{sec:dataset}). With only two auxiliary datasets, both
below the threshold, this result is consistent with low DHS
signalling unsuitable fusion partners, but it cannot establish the
threshold as a predictor of fusion failure.

\subsection{Transformer Model Comparison}

Table~\ref{tab:transformers} reports the results for all four
transformer models on the BanglishDepYT gold test set across
ten random seeds under identical hyperparameters and data splits.
To confirm the reliability of these results, each model was
evaluated across multiple independent experimental runs, and the
findings remained consistent throughout, confirming the numbers
reflect genuine model characteristics rather than experimental
artifacts.

\begin{table}[htbp]
\centering
\caption{Transformer model comparison on BanglishDepYT gold test
set (10 seeds, EPH preprocessing, identical hyperparameters).
$p$-values from paired $t$-test against SahajBERT baseline unless
noted. Bold = best value per column.}
\label{tab:transformers}
\footnotesize
\begin{tabular}{lrrrrl}
\toprule
\textbf{Model}
  & \textbf{F1}$_\mu$
  & \textbf{F1}$_\sigma$
  & \textbf{MCC}$_\mu$
  & \textbf{MCC}$_\sigma$
  & \textbf{$p$-value} \\
\midrule
SahajBERT baseline
  & 0.7805 & 0.0287 & 0.5546 & 0.0584 & --- \\
\textbf{DAPT-SahajBERT}
  & \textbf{0.8161} & \textbf{0.0101}
  & \textbf{0.6272} & 0.0257
  & $\mathbf{0.0013}^{**}$ \\
MuRIL
  & 0.8136 & 0.0110 & 0.6223 & 0.0246
  & 0.589$^\dagger$ \\
IndicBERTv2
  & 0.7933 & 0.0322 & 0.5773 & 0.0657
  & 0.372 \\
\midrule
\multicolumn{6}{l}{\footnotesize
  $^{**}p<0.01$ (paired over 10 training seeds); $d_z=1.450$.
  Test-level McNemar $p=0.170$ (Table~\ref{tab:mcnemar}).}\\
\multicolumn{6}{l}{\footnotesize
  $^\dagger p$-value for DAPT-SahajBERT vs MuRIL comparison
  (not vs baseline).}\\
\bottomrule
\end{tabular}
\end{table}

DAPT-SahajBERT obtains the highest mean weighted F1 (0.8161) and
the smallest standard deviation across ten seeds (0.0101) among the
four transformers. Its MCC of 0.6272 is also the highest,
indicating that the improvement is not driven by majority-class
predictions.

Over the ten training seeds, the improvement over the SahajBERT
baseline is statistically significant ($p=0.0013$, $d_z=1.450$),
and DAPT-SahajBERT has higher F1 on 8 of the 10 seeds
(Table~\ref{tab:perseed}). This shows that the gain is robust to
training randomness on this test set. It does not show that the
gain generalizes: all seeds share the same 150 test comments, the
prediction-level McNemar test is not significant ($p=0.170$), and
the DAPT corpus contained the unlabeled test text
(Section~\ref{sec:dataset}). The result is consistent with a
benefit from domain-adaptive pretraining on informal Bengali text,
but confirming it requires a larger test set and a DAPT corpus with
the gold comments removed.

MuRIL achieves a mean F1 of 0.8136, not significantly different
from DAPT-SahajBERT ($p=0.589$). A non-significant difference is
not proof of equivalence, but the two means differ by only 0.0025. MuRIL was built by
Google using multilingual corpora across 17 Indian languages,
including transliterated text that closely resembles Banglish.
DAPT-SahajBERT starts from a much smaller monolingual model and
adds only 163{,}217 domain-specific unlabeled comments.
That the two models perform similarly despite the large resource
gap between them suggests that targeted domain pretraining on
modest unlabeled data can be a competitive alternative to
large-scale multilingual pretraining for this task, subject to the
test-set caveats above.

IndicBERTv2 does not significantly outperform the SahajBERT
baseline ($p=0.372$, $d_z=0.297$). This negative result is
informative: simply being a larger multilingual model does not
guarantee improvement on informal Bengali depression text.
The relevance of the pretraining data to the target domain
appears to matter more than the breadth of multilingual coverage,
as MuRIL's stronger Indian-language focus gives it an advantage
that IndicBERTv2's broader but less specialized pretraining does not.

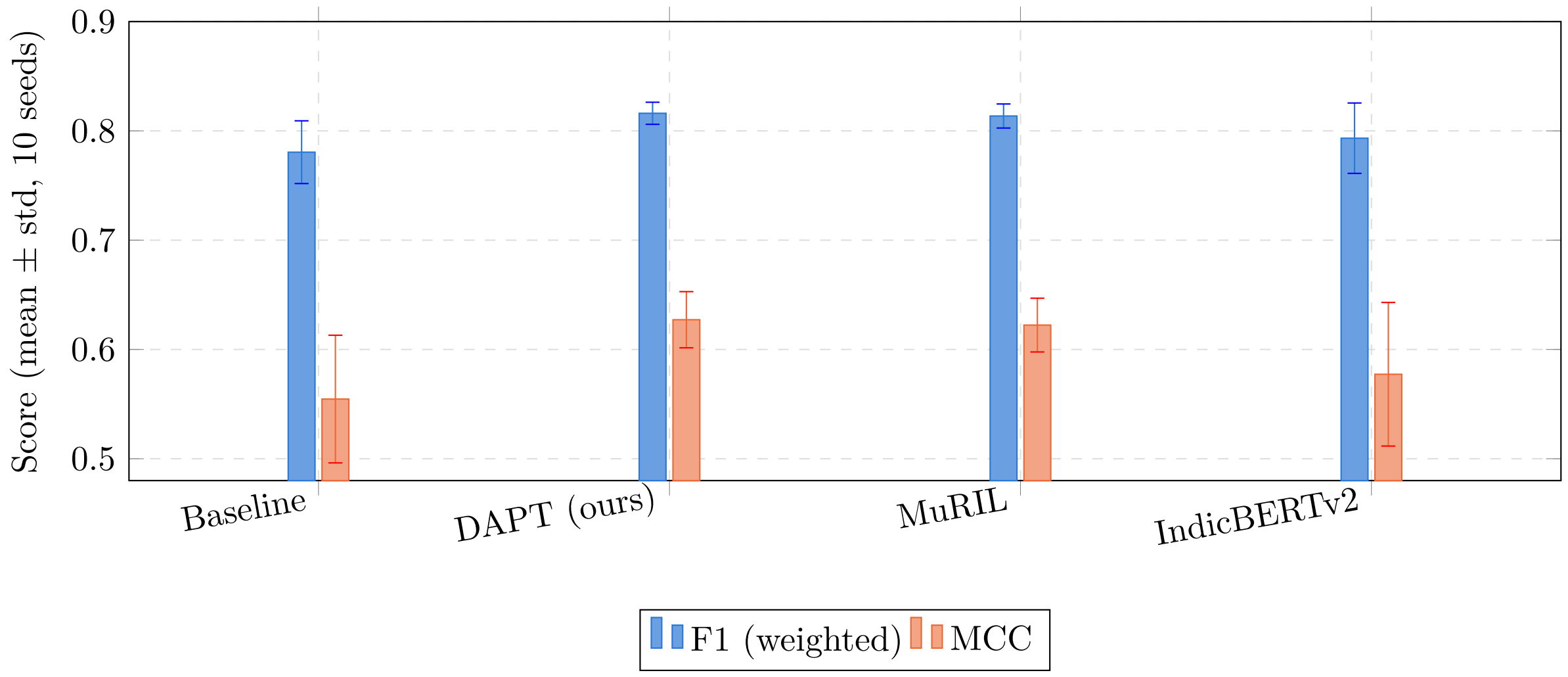
\begin{figure}[htbp]
\centering
\begin{tikzpicture}
\begin{axis}[
  ybar,
  bar width=0.27cm,
  width=\columnwidth,
  height=6.2cm,
  ymin=0.48, ymax=0.90,
  symbolic x coords={
    Baseline, {DAPT (ours)}, MuRIL, IndicBERTv2},
  xtick=data,
  x tick label style={font=\small, rotate=10, anchor=east},
  ylabel={Score (mean $\pm$ std, 10 seeds)},
  ylabel style={font=\small},
  yticklabel style={font=\small,
    /pgf/number format/fixed,
    /pgf/number format/precision=2},
  legend style={
    at={(0.5,-0.28)}, anchor=north,
    legend columns=2, font=\small},
  enlarge x limits=0.18,
  error bars/y dir=both,
  error bars/y explicit,
  grid=major,
  grid style={dashed,gray!25},
]
\addplot+[
  fill=multicol!70, draw=multicol,
  error bars/.cd, y dir=both, y explicit
] coordinates {
  (Baseline,       0.7805) +- (0, 0.0287)
  ({DAPT (ours)},  0.8161) +- (0, 0.0101)
  (MuRIL,          0.8136) +- (0, 0.0110)
  (IndicBERTv2,    0.7933) +- (0, 0.0322)
};
\addplot+[
  fill=ytcol!60, draw=ytcol,
  error bars/.cd, y dir=both, y explicit
] coordinates {
  (Baseline,       0.5546) +- (0, 0.0584)
  ({DAPT (ours)},  0.6272) +- (0, 0.0257)
  (MuRIL,          0.6223) +- (0, 0.0246)
  (IndicBERTv2,    0.5773) +- (0, 0.0657)
};
\legend{F1 (weighted), MCC}
\end{axis}
\end{tikzpicture}
\caption{Transformer model comparison (10 seeds, error bars =
std). DAPT-SahajBERT achieves the best mean F1 and MCC and the
smallest variance across seeds. It is not significantly different
from MuRIL ($p=0.589$) and outperforms the baseline across seeds
($p=0.0013$, $d_z=1.450$). IndicBERTv2 does not
significantly outperform the baseline ($p=0.372$).}
\label{fig:models}
\end{figure}

Figure~\ref{fig:models} visualizes mean F1 and MCC scores with
standard deviation error bars across all four models.

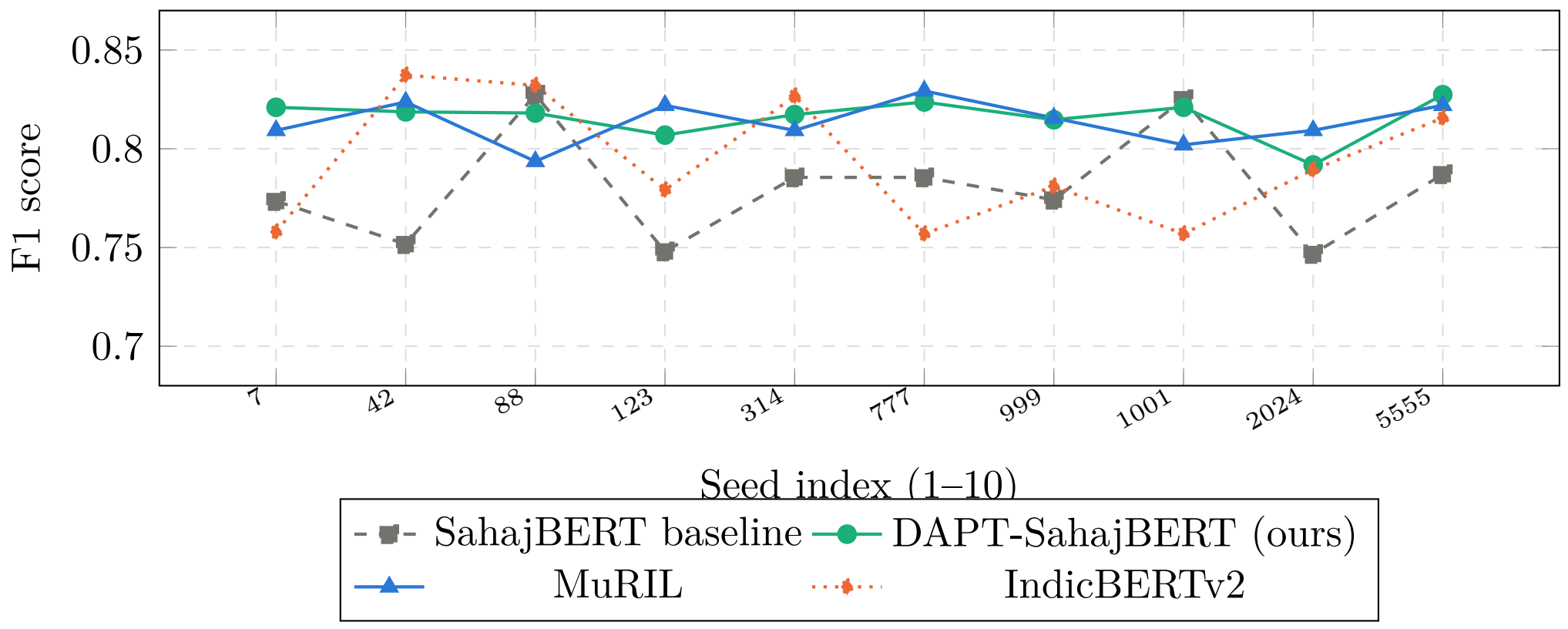
\begin{figure}[htbp]
\centering
\begin{tikzpicture}
\begin{axis}[
  width=0.85\columnwidth,
  height=4.8cm,
  xlabel={Seed index (1--10)},
  ylabel={F1 score},
  xlabel style={font=\small},
  ylabel style={font=\small},
  xticklabel style={font=\small},
  yticklabel style={font=\small,
    /pgf/number format/fixed,
    /pgf/number format/precision=2},
  ymin=0.68, ymax=0.87,
  xtick={1,2,3,4,5,6,7,8,9,10},
  xticklabels={
    7,42,88,123,314,777,999,1001,2024,5555},
  x tick label style={rotate=30, anchor=east, font=\tiny},
  legend style={
    at={(0.5,-0.30)}, anchor=north,
    legend columns=2, font=\small},
  grid=major,
  grid style={dashed,gray!25},
  mark size=2pt,
]
\addplot[
  color=bsmddcol, mark=square*, thick,
  dashed
] coordinates {
  (1,0.7735)(2,0.7517)(3,0.8274)
  (4,0.7480)(5,0.7855)(6,0.7855)
  (7,0.7742)(8,0.8251)(9,0.7467)(10,0.7871)};
\addlegendentry{SahajBERT baseline}
\addplot[
  color=daptcol, mark=*, thick
] coordinates {
  (1,0.8210)(2,0.8187)(3,0.8181)
  (4,0.8069)(5,0.8173)(6,0.8237)
  (7,0.8147)(8,0.8210)(9,0.7918)(10,0.8274)};
\addlegendentry{DAPT-SahajBERT (ours)}
\addplot[
  color=multicol, mark=triangle*, thick
] coordinates {
  (1,0.8093)(2,0.8237)(3,0.7936)
  (4,0.8220)(5,0.8093)(6,0.8293)
  (7,0.8156)(8,0.8020)(9,0.8093)(10,0.8220)};
\addlegendentry{MuRIL}
\addplot[
  color=ytcol, mark=diamond*, thick,
  dotted
] coordinates {
  (1,0.7579)(2,0.8373)(3,0.8321)
  (4,0.7791)(5,0.8267)(6,0.7569)
  (7,0.7810)(8,0.7569)(9,0.7894)(10,0.8156)};
\addlegendentry{IndicBERTv2}
\end{axis}
\end{tikzpicture}
\caption{Per-seed F1 scores for all four transformer models
across ten seeds, in ascending seed order. DAPT-SahajBERT (solid
line) outperforms the baseline (dashed) on 8 of 10 seeds; the
baseline is higher at seeds 88 and 1001. DAPT-SahajBERT maintains
a tighter band than IndicBERTv2 (dotted), reflecting its lower
variance ($\sigma=0.0101$). MuRIL achieves similar performance
to DAPT-SahajBERT throughout.}
\label{fig:perseed}
\end{figure}

\subsection{Does LLM-Assisted Annotation at Scale Actually Help?}

Our Qwen2.5-7B annotator achieved Cohen's $\kappa=0.611$ against
the 1{,}000 human-verified gold labels, confirming substantial
agreement. We then applied it to 101{,}000 comments from the
clean corpus, retaining 96{,}071 high-confidence predictions, of
which 66{,}851 remain after EPH, gold-overlap removal and
deduplication (Section~\ref{sec:dataset}). Semi-supervised training
compares SahajBERT trained on the 700 gold training examples alone
against training on those examples combined with the 66{,}851
pseudo-labeled training examples.
Table~\ref{tab:semisupervised} summarizes the results.

\begin{table}[htbp]
\centering
\caption{Semi-supervised SahajBERT results (5 seeds). Combined
training adds 66{,}851 pseudo-labeled samples to the 700 gold
training examples, producing a 96-fold increase in training
set size at no additional human annotation cost.}
\label{tab:semisupervised}
\footnotesize
\setlength{\tabcolsep}{4pt}
\begin{tabular}{lrrrrl}
\toprule
\textbf{Training data}
  & \textbf{F1}
  & \textbf{F1 std}
  & \textbf{MCC}
  & \textbf{AUC}
  & \textbf{N} \\
\midrule
Gold only
  & 0.7741 & 0.0286 & 0.5395 & 0.8499 & 700 \\
\textbf{Gold + pseudo}
  & \textbf{0.7935} & \textbf{0.0182}
  & \textbf{0.5870} & \textbf{0.8731}
  & 67{,}551 \\
\bottomrule
\multicolumn{6}{l}{\footnotesize
  AUC improvement: $p=0.044^*$.
  F1 and MCC show positive trends.}\\
\end{tabular}
\end{table}

Adding pseudo-labels raises F1 by 0.0194, AUC by 0.0232
($p=0.044$), and MCC by 0.0475. The standard deviation of F1
also drops from 0.0286 to 0.0182, suggesting the larger training
set stabilizes learning as well as improving it. This result
provides a \textbf{reproducible zero-cost annotation scaling
pathway} for mental health NLP in data-scarce languages, directly
applicable to other South Asian and code-mixed settings where
expert annotators are scarce.

\begin{figure}[htbp]
\centering
\begin{tikzpicture}
\begin{axis}[
  ybar,
  bar width=0.55cm,
  width=0.72\columnwidth,
  height=4.6cm,
  ymin=0.45, ymax=0.95,
  symbolic x coords={F1, MCC, AUC},
  xtick=data,
  x tick label style={font=\small},
  ylabel={Score (mean, 5 seeds)},
  ylabel style={font=\small},
  yticklabel style={font=\small,
    /pgf/number format/fixed,
    /pgf/number format/precision=2},
  legend style={
    at={(0.5,-0.22)}, anchor=north,
    legend columns=2, font=\small},
  enlarge x limits=0.30,
  nodes near coords,
  nodes near coords style={
    font=\scriptsize, rotate=90,
    anchor=west, yshift=2pt},
  point meta=explicit symbolic,
  grid=major,
  grid style={dashed,gray!25},
]
\addplot[fill=goldcol!70, draw=goldcol,
         point meta=explicit symbolic]
  coordinates{
    (F1,  0.7741)[0.7741]
    (MCC, 0.5395)[0.5395]
    (AUC, 0.8499)[0.8499]};
\addplot[fill=combinedcol!70, draw=combinedcol,
         point meta=explicit symbolic]
  coordinates{
    (F1,  0.7935)[0.7935]
    (MCC, 0.5870)[0.5870]
    (AUC, 0.8731)[0.8731]};
\legend{Gold only (700 samples),
        Gold + pseudo-labels (67{,}551)}
\end{axis}
\end{tikzpicture}
\caption{Semi-supervised training results (5 seeds). Adding
66{,}851 filtered LLM pseudo-labels ($\kappa=0.611$) improves all
three metrics consistently. AUC improvement is statistically
significant ($p=0.044$). The 96-fold training set increase
comes at no additional human annotation cost.}
\label{fig:semisup}
\end{figure}
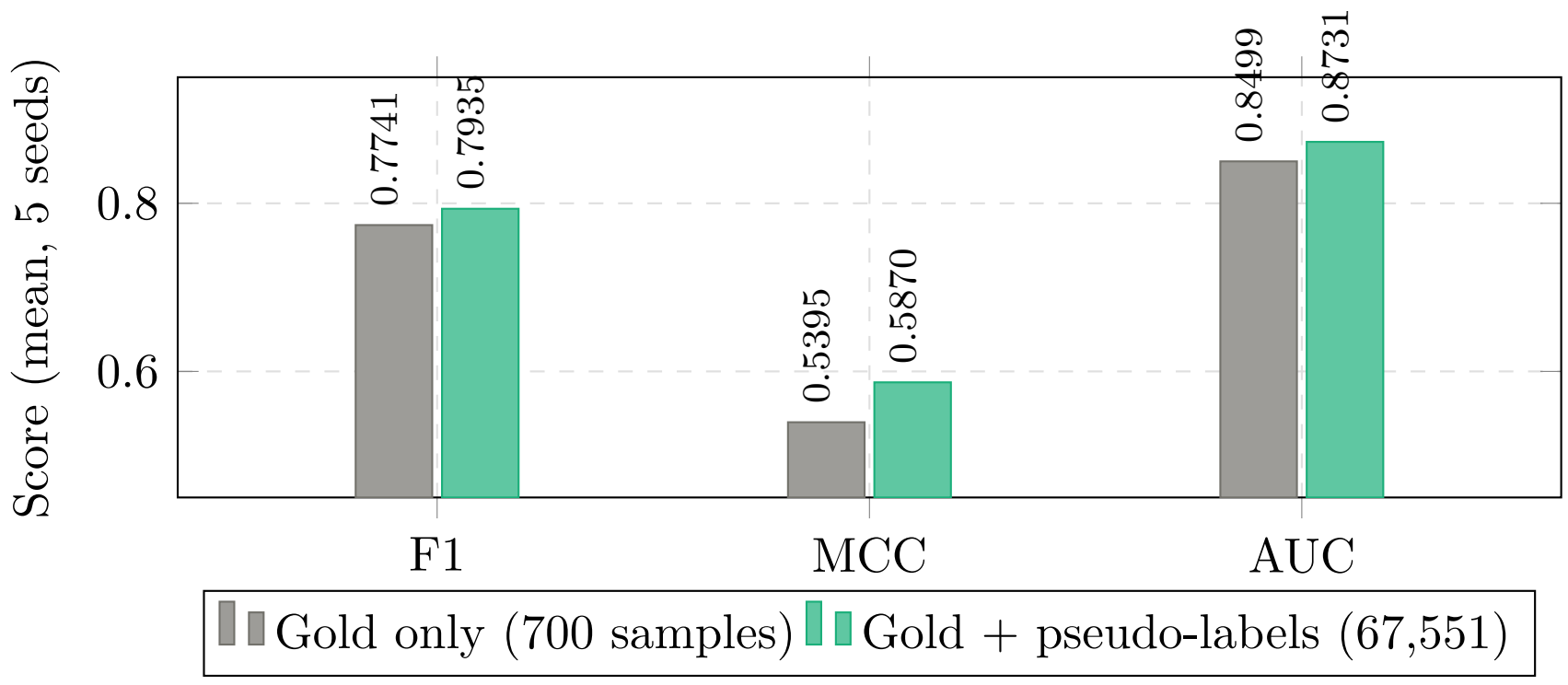

\subsection{How Do All Results Fit Together?}

\begin{figure}[htbp]
\centering
\begin{tikzpicture}
\begin{axis}[
  xbar,
  bar width=0.42cm,
  width=0.72\columnwidth,
  height=5.5cm,
  xmin=0.60, xmax=0.92,
  ymin=0.5, ymax=7.5,
  ytick={1,2,3,4,5,6,7},
  yticklabels={
    {SVM + P2 (standard)},
    {SVM + P4 (EPH)},
    {SahajBERT baseline},
    {Semi-supervised},
    {IndicBERTv2},
    {MuRIL},
    {DAPT-SahajBERT (ours)}
  },
  y tick label style={font=\small},
  x tick label style={font=\small,
    /pgf/number format/fixed,
    /pgf/number format/precision=2},
  xlabel={F1 score},
  xlabel style={font=\small},
  enlarge y limits=0.10,
  nodes near coords,
  nodes near coords style={font=\scriptsize, xshift=2pt},
  point meta=explicit symbolic,
  grid=major,
  grid style={dashed,gray!25},
]
\addplot[fill=bsmddcol!60, draw=gray!60, bar shift=0pt]
  coordinates {(0.7062,1)[0.7062]};
\addplot[fill=ytcol!80, draw=gray!60, bar shift=0pt]
  coordinates {(0.8570,2)[0.8570]};
\addplot[fill=bsmddcol!60, draw=gray!60, bar shift=0pt]
  coordinates {(0.7805,3)[0.7805]};
\addplot[fill=combinedcol!70, draw=gray!60, bar shift=0pt]
  coordinates {(0.7935,4)[0.7935]};
\addplot[fill=ytcol!60, draw=gray!60, bar shift=0pt]
  coordinates {(0.7933,5)[0.7933]};
\addplot[fill=multicol!70, draw=gray!60, bar shift=0pt]
  coordinates {(0.8136,6)[0.8136]};
\addplot[fill=daptcol!80, draw=gray!60, bar shift=0pt]
  coordinates {(0.8161,7)[0.8161]};
\end{axis}
\end{tikzpicture}
\caption{Summary of F1 scores across key experimental
conditions. EPH produces the largest observed improvement
in the SVM preprocessing comparison, increasing F1 from
0.7062 under standard preprocessing (P2) to 0.8570 under
EPH (P4), a gain of 0.1508. DAPT-SahajBERT achieves the
highest mean F1 among the four evaluated transformer models.
Semi-supervised training and IndicBERTv2 obtain scores near
0.793, while MuRIL and DAPT-SahajBERT obtain the highest
mean F1 scores among the transformers. Transformer results
are means across 10 seeds; SVM results are deterministic;
semi-supervised results are means across 5 seeds. The two SVM bars
are measured on the 126-comment test split of the balanced subset;
all other bars use the 150-comment gold test split, so the SVM bars
are not directly comparable with the others. On the balanced
subset, BanglaBERT reaches 0.8621 with EPH
(Table~\ref{tab:yt_allmodels}).}
\label{fig:summary}
\end{figure}
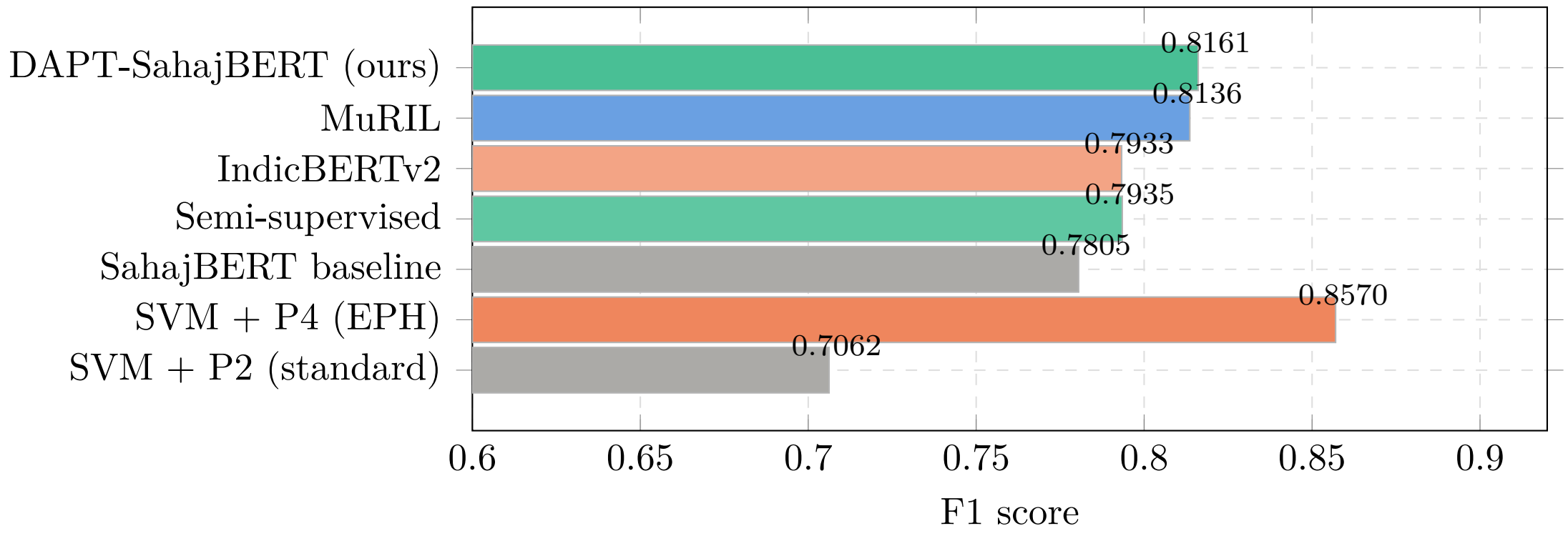

Figure~\ref{fig:summary} brings together the key results from
the different experimental conditions, providing an overview
of the F1 scores achieved by the configurations evaluated in
this paper.

Three conclusions follow from these results. First, preprocessing
choice can substantially affect classification performance on
native informal Bengali text, with EPH yielding a 0.1508 F1
improvement over standard preprocessing for the SVM and 0.0266 for
BanglaBERT, but no gain for SahajBERT or the BiLSTM. Second, among
the four evaluated transformer models, DAPT-SahajBERT achieves
the highest mean F1 (0.8161) and mean MCC (0.6272), while
exhibiting lower variation across random seeds than the
SahajBERT baseline. This is consistent with a benefit from
domain-adaptive pretraining, subject to the test-set size and
contamination caveats discussed above.
Third, LLM-assisted annotation offers a scalable way to expand
the training data, with semi-supervised training improving
classification performance over gold-only training. Together,
these findings highlight the complementary contributions of
emotion-preserving preprocessing, domain-adaptive pretraining,
and LLM-assisted pseudo-labeling.

\section{Statistical Analysis}
\label{sec:stats}

\subsection{Why Statistics Matter Here}

Reporting a single F1 score and calling it a result is not enough
for a paper making comparative claims. Without running the same
experiment multiple times and testing whether the differences
between models are larger than the variation between runs, we
cannot tell whether our findings reflect genuine model
characteristics or initialization artifacts. This section presents
the statistical evidence for the comparisons in
Section~\ref{sec:results}. Seed-level tests address robustness to
training randomness; the McNemar test addresses differences at the
level of individual test predictions.

\subsection{DAPT vs Baseline: The Central Comparison}

The most important comparison in this paper is between SahajBERT
fine-tuned directly on the 700-sample gold training set and
DAPT-SahajBERT, which first read all 163{,}217 unlabeled comments
before fine-tuning on the same training split. We ran both models
ten times with different random seeds and collected F1 from each
run. Table~\ref{tab:perseed} lists the paired results.

\begin{table}[H]
\centering
\caption{Per-seed F1 scores for SahajBERT baseline and
DAPT-SahajBERT across ten seeds. DAPT-SahajBERT has the higher
F1 on 8 of 10 seeds.}
\label{tab:perseed}
\footnotesize
\setlength{\tabcolsep}{4pt}
\begin{tabular}{lrrl}
\toprule
\textbf{Seed}
  & \textbf{Baseline F1}
  & \textbf{DAPT F1}
  & \textbf{Outcome} \\
\midrule
7    & 0.7735 & 0.8210 & DAPT wins \\
42   & 0.7517 & 0.8187 & DAPT wins \\
88   & 0.8274 & 0.8181 & Baseline wins \\
123  & 0.7480 & 0.8069 & DAPT wins \\
314  & 0.7855 & 0.8173 & DAPT wins \\
777  & 0.7855 & 0.8237 & DAPT wins \\
999  & 0.7742 & 0.8147 & DAPT wins \\
1001 & 0.8251 & 0.8210 & Baseline wins \\
2024 & 0.7467 & 0.7918 & DAPT wins \\
5555 & 0.7871 & 0.8274 & DAPT wins \\
\midrule
\textbf{Mean} & 0.7805 & 0.8161 & --- \\
\textbf{Std}  & 0.0287 & 0.0101 & --- \\
\bottomrule
\multicolumn{4}{l}{\footnotesize
  Paired $t$-test: $t=4.586$, $p=0.0013$; $d_z=1.450$;}\\
\multicolumn{4}{l}{\footnotesize
  Wilcoxon signed-rank $p=0.0098$.}\\
\end{tabular}
\end{table}

DAPT-SahajBERT has the higher F1 on 8 of the 10 seeds; the
baseline is higher at seeds 88 and 1001, where it reaches its two
best scores (0.8274 and 0.8251). A paired $t$-test gives $t=4.586$,
$p=0.0013$, $d_z=1.450$, and the non-parametric Wilcoxon
signed-rank test agrees ($p=0.0098$). The standard deviation drops
from 0.0287 (baseline) to 0.0101 (DAPT), a 65\% reduction, so DAPT
also makes fine-tuning less sensitive to the random seed. These
tests concern training randomness on a single 150-comment test
set; Section~\ref{sec:mcnemar} reports the corresponding
prediction-level test.

\begin{figure}[H]
\centering
\begin{tikzpicture}
\begin{axis}[
  width=0.85\columnwidth,
  height=4.2cm,
  ymin=0.70, ymax=0.86,
  xlabel={Seed index},
  ylabel={F1 score},
  xlabel style={font=\small},
  ylabel style={font=\small},
  xticklabel style={font=\small},
  yticklabel style={font=\small,
    /pgf/number format/fixed,
    /pgf/number format/precision=2},
  xtick={1,2,3,4,5,6,7,8,9,10},
  xticklabels={7,42,88,123,314,777,999,1001,2024,5555},
  x tick label style={
    rotate=30, anchor=east, font=\tiny},
  legend style={
    at={(0.98,0.05)}, anchor=south east,
    font=\small},
  grid=major,
  grid style={dashed,gray!25},
  mark size=2.5pt,
]

\addplot[color=bsmddcol, mark=square*, thick, dashed] coordinates {
  (1,0.7735)(2,0.7517)(3,0.8274)
  (4,0.7480)(5,0.7855)(6,0.7855)
  (7,0.7742)(8,0.8251)(9,0.7467)(10,0.7871)};
\addlegendentry{Baseline}
\addplot[color=daptcol, mark=*, thick] coordinates {
  (1,0.8210)(2,0.8187)(3,0.8181)
  (4,0.8069)(5,0.8173)(6,0.8237)
  (7,0.8147)(8,0.8210)(9,0.7918)(10,0.8274)};
\addlegendentry{DAPT-SahajBERT}
\addplot[fill=daptcol!15, draw=none, forget plot] coordinates {
  (1,0.8210)(2,0.8187)(3,0.8181)
  (4,0.8069)(5,0.8173)(6,0.8237)
  (7,0.8147)(8,0.8210)(9,0.7918)(10,0.8274)
  (10,0.7871)(9,0.7467)(8,0.8251)
  (7,0.7742)(6,0.7855)(5,0.7855)
  (4,0.7480)(3,0.8274)(2,0.7517)(1,0.7735)};
\end{axis}
\end{tikzpicture}
\caption{Paired per-seed F1 comparison between SahajBERT
baseline (dashed) and DAPT-SahajBERT (solid). The shaded
region shows the per-seed difference. DAPT-SahajBERT is higher
on 8 of 10 seeds; the baseline is higher at seeds 88 and 1001.
Paired $t$-test: $p=0.0013$, $d_z=1.450$.}
\label{fig:paired_seeds}
\end{figure}
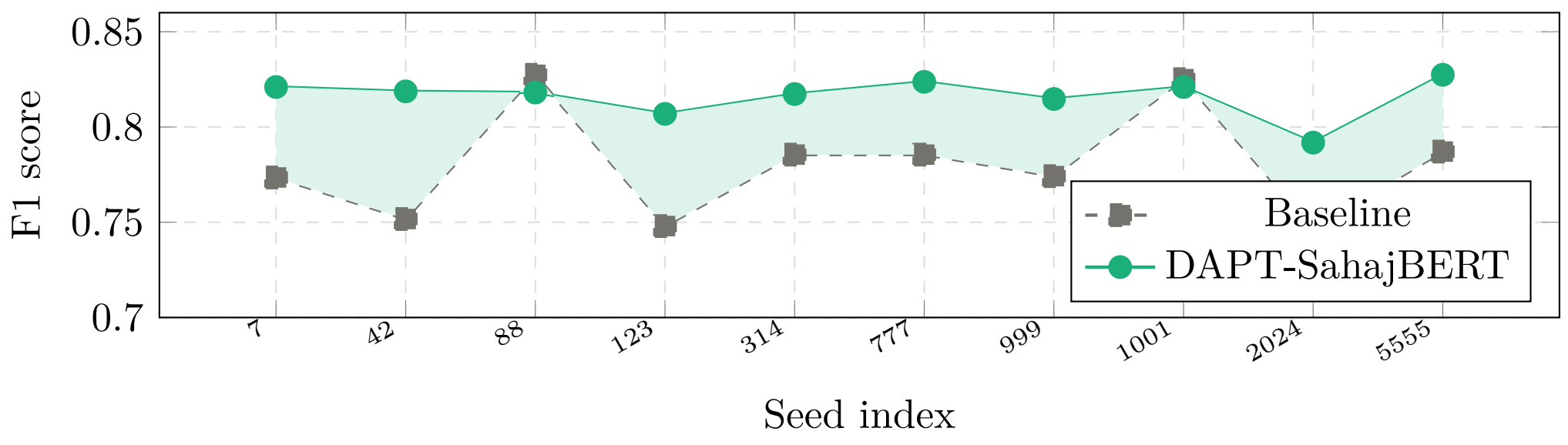

\subsection{McNemar Test: What Kind of Errors Change?}
\label{sec:mcnemar}

The paired $t$-test tells us overall F1 improves. McNemar's test
tells us whether DAPT actually corrects errors the baseline makes,
or merely shuffles which examples are classified correctly.
Table~\ref{tab:mcnemar} reports the paired prediction counts at
seed 42 on the 150-sample gold test set.

\begin{table}[htbp]
\centering
\caption{McNemar contingency table: baseline vs
DAPT-SahajBERT predictions on 150 gold test samples
(seed 42). DAPT corrects nearly twice as many errors
as it introduces.}
\label{tab:mcnemar}
\begin{tabular}{lrr}
\toprule
  & \textbf{DAPT correct} & \textbf{DAPT wrong} \\
\midrule
\textbf{Baseline correct} & 106 &  9 \\
\textbf{Baseline wrong}   &  17 & 18 \\
\bottomrule
\multicolumn{3}{l}{\footnotesize
  McNemar $p=0.170$ (not significant);}\\
\multicolumn{3}{l}{\footnotesize
  DAPT corrects 1.9$\times$ more errors than it introduces.}\\
\end{tabular}
\end{table}

DAPT corrects 17 examples the baseline gets wrong, while the
baseline corrects only 9 that DAPT gets wrong --- a ratio of
nearly 2:1 in DAPT's favour, but only 26 of the 150 test comments
are classified differently by the two models. McNemar's test does
not reach significance ($p=0.170$). This is the most direct test
of whether DAPT improves predictions on test data, and on this
test set it does not establish an improvement. The seed-level
result above shows that the direction of the difference is stable
across training runs, not that it would hold on new data. A
larger held-out test set, or cross-validation with bootstrap
confidence intervals on test predictions, is needed to settle
this.

\subsection{All Model Pairwise Comparisons}

Table~\ref{tab:pairwise} reports all pairwise paired
$t$-tests across the four transformer models.

\begin{table}[htbp]
\centering
\caption{Pairwise statistical comparisons across four transformer
models (paired $t$-test, 10 seeds). Only comparisons involving
the baseline are tested against it; the DAPT vs multilingual
comparisons are cross-group.}
\label{tab:pairwise}
\resizebox{\columnwidth}{!}{%
\begin{tabular}{lrrrrrl}
\toprule
\textbf{Comparison}
  & $\Delta$\textbf{F1}
  & \textbf{$t$}
  & \textbf{$p$}
  & \textbf{$p_{\text{Holm}}$}
  & \textbf{$d_z$}
  & \textbf{Result (Holm)} \\
\midrule
Baseline $\to$ DAPT
  & $+0.0356$ & 4.586 & $0.0013$ & $\mathbf{0.008}^{**}$
  & 1.450 & Significant \\
Baseline $\to$ MuRIL
  & $+0.0331$ & 2.874 & $0.018$ & 0.092
  & 0.909 & Not significant \\
Baseline vs IndicBERTv2
  & $+0.0128$ & 0.939 & 0.372 & 0.744
  & 0.297 & Not significant \\
DAPT vs MuRIL
  & $-0.0025$ & $-0.560$ & 0.589 & 0.744
  & $-0.177$ & Not significant \\
DAPT vs IndicBERTv2
  & $-0.0228$ & $-2.155$ & 0.060 & 0.238
  & $-0.681$ & Not significant \\
MuRIL vs IndicBERTv2
  & $-0.0203$ & $-1.819$ & 0.102 & 0.307
  & $-0.575$ & Not significant \\
\bottomrule
\multicolumn{7}{l}{\footnotesize
  $^{**}p_{\text{Holm}}<0.01$. $p_{\text{Holm}}$: Holm correction
  over the six comparisons. All tests are paired over 10 training
  seeds on one test set.}\\
\end{tabular}%
}
\end{table}

Without correction, both DAPT-SahajBERT and MuRIL outperform the
baseline at $p<0.05$ and IndicBERTv2 does not. After Holm
correction over the six comparisons, only the DAPT-SahajBERT
versus baseline difference remains significant ($p_{\text{Holm}}
=0.008$); the MuRIL versus baseline difference does not
($p_{\text{Holm}}=0.092$). DAPT-SahajBERT and MuRIL are not
significantly different.

\subsection{Semi-Supervised Training Significance}

Table~\ref{tab:ss_significance} reports paired $t$-tests for
the semi-supervised comparison across five seeds.

\begin{table}[htbp]
\centering
\caption{Semi-supervised significance tests (paired $t$-test,
5 seeds). AUC improvement is statistically significant.
F1 and MCC show consistent positive trends.}
\label{tab:ss_significance}
\begin{tabular}{lrrrr}
\toprule
\textbf{Metric}
  & \textbf{Gold only}
  & \textbf{Combined}
  & $\Delta$
  & \textbf{$p$} \\
\midrule
F1  & 0.7741 & 0.7935 & $+0.0194$ & 0.089 \\
MCC & 0.5395 & 0.5870 & $+0.0475$ & 0.071 \\
AUC & 0.8499 & 0.8731 & $+0.0232$ & $\mathbf{0.044}^{*}$ \\
\bottomrule
\multicolumn{5}{l}{\footnotesize $^{*}p<0.05$.
  F1 and MCC do not reach significance at $n=5$ seeds;}\\
\multicolumn{5}{l}{\footnotesize the positive trend is
  consistent across all five seeds for all three metrics.}\\
\end{tabular}
\end{table}

\subsection{Model Stability}

Table~\ref{tab:stability} compares model stability using the
coefficient of variation (CV) across ten seeds.

\begin{table}[htbp]
\centering
\caption{Model stability across ten seeds. CV = std / mean.
Lower CV indicates more reliable deployment behaviour.
DAPT-SahajBERT achieves the best performance and the
lowest variance simultaneously.}
\label{tab:stability}
\begin{tabular}{lrrr}
\toprule
\textbf{Model}
  & \textbf{F1 mean}
  & \textbf{F1 std}
  & \textbf{CV (\%)} \\
\midrule
SahajBERT baseline & 0.7805 & 0.0287 & 3.68 \\
\textbf{DAPT-SahajBERT}
  & \textbf{0.8161} & \textbf{0.0101} & \textbf{1.24} \\
MuRIL              & 0.8136 & 0.0110 & 1.35 \\
IndicBERTv2        & 0.7933 & 0.0322 & 4.06 \\
\bottomrule
\end{tabular}
\end{table}

\begin{figure}[htbp]
\centering
\begin{tikzpicture}
\begin{axis}[
  ybar,
  bar width=0.65cm,
  width=0.80\columnwidth,
  height=5.0cm,
  ymin=0, ymax=5.0,
  symbolic x coords={
    Baseline, {DAPT (ours)}, MuRIL, IndicBERTv2},
  xtick=data,
  x tick label style={font=\small, rotate=10, anchor=east},
  ylabel={CV (\%)},
  ylabel style={font=\small},
  yticklabel style={font=\small},
  nodes near coords,
  nodes near coords style={font=\small, yshift=2pt},
  enlarge x limits=0.20,
  grid=major,
  grid style={dashed,gray!25},
]
\addplot[
  fill=multicol!60,
  draw=gray!60
] coordinates {
  (Baseline,       3.68)
  ({DAPT (ours)},  1.24)
  (MuRIL,          1.35)
  (IndicBERTv2,    4.06)
};
\end{axis}
\end{tikzpicture}
\caption{Coefficient of variation (CV) across ten seeds for
each transformer model. Lower CV means less sensitivity to the
random seed. DAPT-SahajBERT achieves the best mean F1 and the
lowest CV simultaneously.}
\label{fig:cv}
\end{figure}
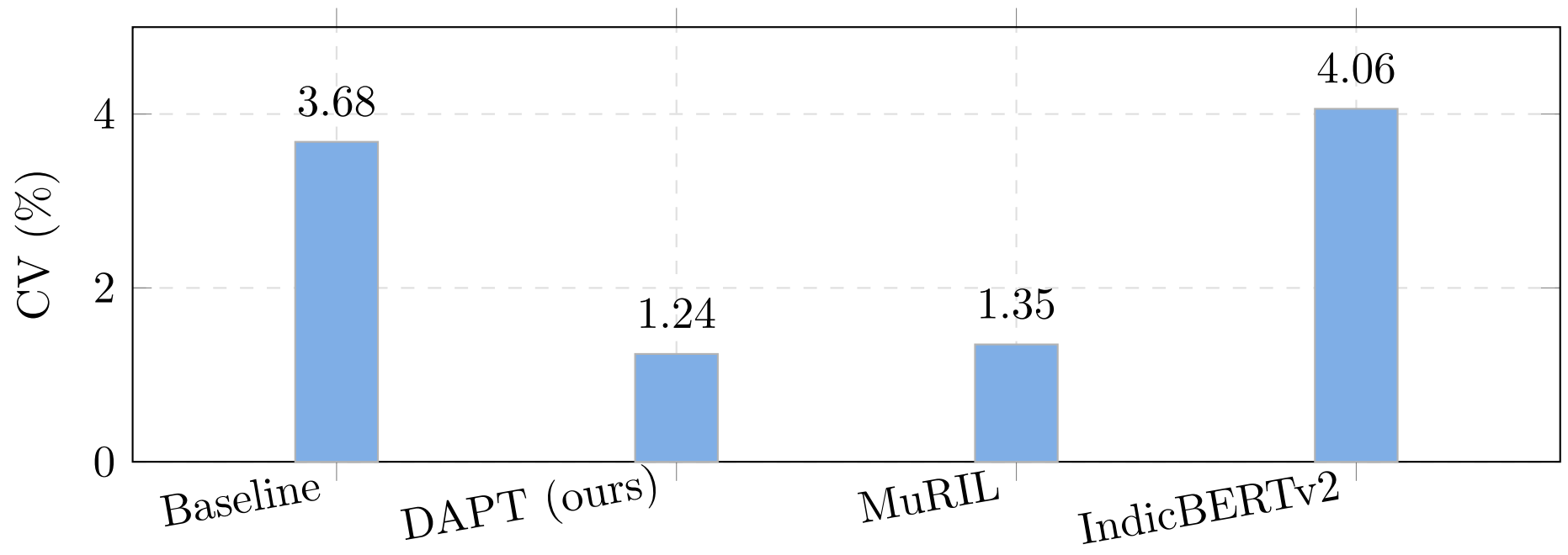

DAPT-SahajBERT has the lowest CV at 1.24\%, meaning it is the
most reliable model in addition to achieving the highest F1 and
MCC. IndicBERTv2 has the highest CV at 4.06\%, nearly three
times more variable than DAPT-SahajBERT, suggesting it is
sensitive to random initialization on this dataset.

\subsection{Error Analysis}

A qualitative examination of misclassified comments highlighted
several challenges in detecting depression-related language in
informal Bengali social media text.

False-positive errors may arise when comments express empathy
toward another person, describe emotional reactions to sad songs
or videos, or use hyperbolic expressions of frustration that do
not necessarily indicate personal depressive symptoms.
False-negative errors may occur when distress is expressed
indirectly or understated, or when comments rely heavily on
Banglish expressions that the model does not adequately capture.

These error patterns suggest two directions for future
improvement: expanding annotated training data to cover empathy,
hyperbole, and understated distress, and improving language-model
coverage of Banglish expressions. Further systematic error
analysis on a clearly specified test set would help quantify the
frequency of these error types.


\section{Limitations and Ethical Considerations}

\subsection{Limitations}

\textbf{The labeled set and test set are small.}
All transformer comparisons rest on a single 150-comment test set.
Running ten seeds measures sensitivity to training randomness, not
whether results hold on new data; at the level of individual test
predictions, the DAPT versus baseline difference is not
significant (McNemar $p=0.170$). The DAPT and model-comparison
results should therefore be treated as preliminary until they are
replicated on a substantially larger gold set (several thousand
comments) or with stratified $k$-fold cross-validation and
bootstrap confidence intervals on test predictions. We release the
full 163{,}217-comment cleaned corpus so that annotation can be
expanded without repeating the collection effort.

\textbf{DAPT corpus overlaps the test text.}
The DAPT corpus contains the unlabeled text of all 1{,}000 gold
comments, including the 150 test comments
(Section~\ref{sec:dataset}). No labels were used during
pretraining, but this exposure may inflate DAPT-SahajBERT's
results relative to the other transformers. Removing the gold
comments from the DAPT corpus and repeating pretraining and
fine-tuning is required to rule this out.

\textbf{The annotation sample is enriched for EPH features.}
Seven hundred of the 1{,}000 gold comments were selected by a score
rewarding emoticons, Banglish, elongation and negation, the same
features EPH preserves. Feature rates in the gold set are therefore
much higher than in the full corpus (Section~\ref{sec:annotation}),
and EPH's benefit on the gold set may overstate its benefit on
randomly sampled comments. Evaluating EPH on the 300 randomly
sampled gold comments alone would address this.

\textbf{No human inter-annotator agreement.}
Annotation was performed by two bilingual annotators, both authors
of this paper, with the involvement of an author trained in
educational and counselling psychology, and each label was reached through discussion until
consensus. We did not
record initial disagreements before discussion, so we cannot report
human inter-annotator agreement. We report human--LLM agreement
($\kappa = 0.611$) but not human--human agreement. Future releases
should have annotators label independently, report agreement
before adjudication, and ground the guideline formally in clinical
criteria such as PHQ-9 symptoms.

\textbf{YouTube is not all of Bengali social media.}
Our corpus reflects a particular kind of Bengali speaker: one who
watches YouTube, engages with mental health content, and comments
publicly. It may underrepresent older speakers, rural communities,
Facebook users, and people whose depression is severe enough that
they do not engage with online content at all. A model trained on
our data may not generalize well to other platforms or demographics
without additional data collection.

\textbf{IndicBERTv2 performance was not significantly above
baseline.}
While we expected all multilingual models to outperform the
monolingual baseline, IndicBERTv2 did not reach significance
($p=0.372$, $d_z=0.297$). This result suggests that broader
multilingual coverage does not automatically improve performance on
informal Bengali text; domain-relevant pretraining data matters
more than the number of languages covered. This is a limitation
of the current model selection and a direction for future
investigation.

\textbf{Tentative labels carry noise.}
Non-depressive training examples were collected from videos about
happiness, motivation, and entertainment, based on the assumption
that comments under these videos are more likely to be
non-depressive. This is reasonable but not guaranteed. The manual
annotation of 1{,}000 samples and LLM annotation with confidence
thresholding reduce but do not fully eliminate this noise.

\textbf{The LLM annotator is not perfect.}
Qwen2.5-7B achieved $\kappa = 0.611$ against human labels, meaning
roughly one in five labels may disagree with a careful human
annotator. We kept only high-confidence predictions to mitigate
this, but residual noise from LLM annotation remains in the
combined training data. The F1 and MCC improvements from
semi-supervised training do not individually reach significance at
$p < 0.05$ with five seeds, though AUC does ($p=0.044$) and all
three metrics show consistent positive trends across all seeds.

\textbf{Explainability analysis is limited.}
We report error patterns from manual examination of 50 misclassified
examples. Formal SHAP or attention visualization analysis, which
would give a more systematic account of which tokens most influence
model predictions, remains as future work.

\subsection{Ethical Considerations}

\textbf{Depression is a sensitive subject.}
Every comment in our dataset was written by a real person, possibly
at a moment of genuine distress. We do not quote individual comments
beyond anonymized examples, we do not attempt to identify any
comment author, and we release the dataset under a usage license
restricted to research purposes.

\textbf{YouTube Terms of Service.}
We collected comments using the official YouTube Data API v3 within
the rate limits and terms of service specified by Google. Comment
text accessed through the public API is intended for public access
by design. Author channel identifiers returned by the API are not
included in the released dataset.

\textbf{The model should not be used for autonomous diagnosis.}
A depression detection model trained on social media comments is not
a clinical diagnostic tool. It may serve as a signal in a larger
support system, such as flagging comments for review by a mental
health professional. Deploying any version of this model to
automatically label individuals as depressed, make clinical
decisions, or take actions without human oversight would be
inappropriate and potentially harmful. We explicitly discourage
such use.

\textbf{Class imbalance reflects reality, not bias.}
Our gold labeled set contains 417 depressive and 583 non-depressive
samples. We did not artificially balance these labels because the
natural imbalance reflects the real distribution of comment types
in Bengali mental health comment sections. We handle imbalance
through weighted loss during training rather than data manipulation.

\textbf{Bengali mental health NLP needs more investment.}
Depression affects millions of Bengali speakers, and the NLP tools
available to serve this population are far behind what exists for
high-resource languages. Every limitation described above reflects
the resource gap that characterizes low-resource mental health NLP
more broadly. We hope this paper and dataset lower the barrier for
future work rather than representing its ceiling.

\section{Conclusion}

This paper set out to answer a question that Bengali mental health
NLP has not seriously asked before: what happens when depression
detection models are tested on text that resembles how Bengali
speakers actually write online, rather than text translated from
English and cleaned before training?

We introduced BanglishDepYT, a Bengali depression dataset collected
natively from YouTube mental health comment sections, comprising
191{,}000 unlabeled comments and 1{,}000 samples verified through
dual bilingual annotation. We showed that this data contains
informal linguistic features that are absent from the translated
datasets examined in our study, including emoticons, Banglish
code-mixing, and emotional elongation, which occur in 22.4\%,
7.9\%, and 16.6\% of comments in the full clean corpus,
respectively.

We proposed EPH preprocessing, which treats these informal
features as signals to preserve rather than noise to remove.
Through the CESS framework, we showed that standard preprocessing
reduces the Banglish Survival Rate of native Bengali social media
text to zero, whereas EPH maintains near-complete signal
preservation on the fused native-text corpus
($\text{CESS}=0.9996$ versus $0.6694$). In the BanglishDepYT
preprocessing experiment, SVM F1 increased from 0.7062 under
standard preprocessing to 0.8570 under EPH, a gain of 0.1508, and
BanglaBERT F1 increased from 0.8355 to 0.8621. SahajBERT and the
BiLSTM did not benefit. The effect of preprocessing is therefore
large for bag-of-words models and model-dependent for pretrained
transformers.

We also investigated domain-adaptive pretraining by continuing
SahajBERT's pretraining on 163{,}217 clean unlabeled YouTube
comments before fine-tuning. Among the four evaluated transformer
models, DAPT-SahajBERT achieved the highest mean F1 (0.8161) and
mean MCC (0.6272). Across ten training seeds it outperformed the
SahajBERT baseline in F1, improving from 0.7805 to 0.8161
($p=0.0013$, $d_z=1.450$; higher on 8 of 10 seeds), and its F1
standard deviation decreased from 0.0287 to 0.0101. At the level of
individual test predictions the difference was not significant
(McNemar $p=0.170$), and the DAPT corpus contained the unlabeled
test text, so this result needs confirmation on a larger,
uncontaminated test set. DAPT-SahajBERT
was not significantly different from MuRIL in the reported
comparison ($p=0.589$). These findings support targeted domain
pretraining as a competitive approach for informal Bengali mental
health text classification without requiring the large-scale
multilingual pretraining used by some alternative models.
IndicBERTv2 did not significantly outperform the SahajBERT
baseline ($p=0.372$), suggesting that broader multilingual
coverage alone does not guarantee better performance on this task.

We further showed that LLM-assisted annotation using a locally
deployed Qwen2.5-7B model achieved substantial agreement with
human labels ($\kappa=0.611$). Retaining 96{,}071 high-confidence
pseudo-labels enabled us to investigate semi-supervised training
as a way to expand the available supervision. The evaluated
semi-supervised configuration improved AUC from 0.8499 to
0.8731 ($p=0.044$), illustrating the potential of pseudo-labeling
to reduce reliance on additional manual annotation.

Finally, our preprocessing experiments revealed that BSMDD
produced nearly identical SVM F1 scores across all five
preprocessing conditions, with a range of only 0.0007. This
finding is consistent with the dataset having been released in
pre-processed form and highlights a limitation of using such
data to assess preprocessing methods. The differences observed
between BSMDD and our native YouTube corpus further emphasize
the need to evaluate Bengali depression detection systems on
informal text that reflects real-world language use.

Overall, this work makes three complementary contributions:
a natively collected Bengali depression dataset, an
emotion-preserving preprocessing method, and an investigation
of domain-adaptive pretraining and LLM-assisted pseudo-labeling.
The results show that preserving informal linguistic signals
and adapting models to the target domain are promising strategies
for low-resource Bengali mental health NLP. All code, models,
and datasets are publicly released to support reproducibility
and further research.

Table~\ref{tab:findings} summarizes the nine principal findings
and their quantitative evidence.

\begin{table}[htbp]
\centering
\caption{Summary of nine principal findings with quantitative
evidence. All numbers are from the reported experimental runs.}
\label{tab:findings}
\footnotesize
\setlength{\tabcolsep}{3pt}
\begin{tabularx}{\columnwidth}{@{}cX@{}}
\toprule
\textbf{ID} & \textbf{Finding} \\
\midrule
F1 & On BanglishDepYT, EPH raised SVM F1 from 0.7062 (P2) to
     0.8570 (P4), +0.1508, and BanglaBERT F1 by +0.0266; SahajBERT
     and BiLSTM did not benefit. \\
F2 & Aggressive cleaning (P3) stayed within 0.01 F1 of standard
     cleaning (P2) in all twelve model--corpus combinations: it
     neither recovered nor substantially worsened the loss from
     P2. \\
F3 & Under P0--P3, YouTube SVM F1 trailed BSMDD by
     0.0950--0.1345; P4 narrowed the gap to 0.0166. \\
F4 & Fusing BSMDD with SentNoB and BanglaHate lowered F1 in all 20
     model--condition pairs (DFG from $-0.0246$ to $-0.0464$). \\
F5 & Both auxiliary datasets had DHS $<0.65$ (0.29, 0.33) and
     both reduced performance when fused; two datasets are too few
     to validate DHS as a predictor. \\
F6 & DAPT-SahajBERT: mean F1 0.8161 versus baseline 0.7805
     over 10 seeds ($p=0.0013$, $d_z=1.450$; 8 of 10 seeds);
     McNemar $p=0.170$ on the 150-comment test set. \\
F7 & DAPT-SahajBERT and MuRIL: mean F1 0.8161 versus 0.8136
     ($p=0.589$; no statistically significant difference
     detected). \\
F8 & BSMDD F1 varied by only 0.0007 across P0--P4, consistent
     with the dataset being released in pre-processed form. \\
F9 & Qwen2.5-7B achieved $\kappa=0.611$; semi-supervised
     training improved AUC from 0.8499 to 0.8731
     ($p=0.044$). \\
\bottomrule
\end{tabularx}
\end{table}

\subsection{Future Work}

The most pressing need is expanded annotation. The 191{,}000
unlabeled comments in BanglishDepYT provide material for
developing a substantially larger gold-standard dataset.
A dataset with 10{,}000 dual-verified labels could support
multi-class severity classification and more extensive
benchmark comparisons.

Combining domain-adaptive pretraining with pseudo-labeled
fine-tuning is a natural next experiment. Our DAPT and
semi-supervised experiments were conducted separately.
Combining DAPT with pseudo-labeled training could help determine
whether their benefits are complementary; the resulting
performance improvement remains to be evaluated.

A systematic investigation of multilingual Indic models with
different pretraining compositions would clarify why MuRIL
outperforms the SahajBERT baseline (significant before, but not
after, multiple-comparison correction) while
IndicBERTv2 does not, and identify which pretraining
characteristics transfer most effectively to informal Bengali
depression text.

Extending EPH and the CESS framework to other South Asian
code-mixed languages, including Hindi, Hinglish, Tamil, and
Sinhala, would broaden the scope of this work and test whether
the observed benefits generalize to other informal social media
settings.

Finally, clinical validation remains essential before responsible
deployment. Collaboration with mental health professionals is
needed to assess model predictions against clinically informed
judgments and to establish the limitations of using social media
text for mental health screening.

\bibliography{references}

\begin{thebibliography}{26}
\providecommand{\natexlab}[1]{#1}
\providecommand{\url}[1]{\texttt{#1}}
\expandafter\ifx\csname urlstyle\endcsname\relax
  \providecommand{\doi}[1]{doi: #1}\else
  \providecommand{\doi}{doi: \begingroup \urlstyle{rm}\Url}\fi

\bibitem[Al-Mahrooqi et~al.(2026)]{mhg2025}
Ahmed Al-Mahrooqi et~al.
\newblock {MentalMARBERT}: Domain-adaptive pre-training and two-stage
  fine-tuning for {Arabic} mental health disorders detection.
\newblock \emph{arXiv preprint arXiv:2606.12649}, 2026.
\newblock URL \url{https://arxiv.org/abs/2606.12649}.

\bibitem[Bhattacharjee et~al.(2022)Bhattacharjee, Hasan, Ahmad, Samin, Islam,
  Iqbal, Rahman, and Shahriyar]{bhattacharjee2022}
Abhik Bhattacharjee, Tahmid Hasan, Wasi~Uddin Ahmad, Kazi~Samin Samin,
  Md~Saiful Islam, Anindya Iqbal, M~Sohel Rahman, and Rifat Shahriyar.
\newblock {BanglaBERT}: Language model pretraining and benchmarks for
  low-resource language understanding evaluation in {Bangla}.
\newblock In \emph{Findings of the Association for Computational Linguistics:
  {NAACL} 2022}, pages 1318--1327. Association for Computational Linguistics,
  2022.
\newblock URL \url{https://aclanthology.org/2022.findings-naacl.98}.

\bibitem[Bojanowski et~al.(2017)Bojanowski, Grave, Joulin, and
  Mikolov]{fasttext2017}
Piotr Bojanowski, Edouard Grave, Armand Joulin, and Tomas Mikolov.
\newblock Enriching word vectors with subword information.
\newblock \emph{Transactions of the Association for Computational Linguistics},
  5:\penalty0 135--146, 2017.

\bibitem[Chowdhury et~al.(2024)]{chowdhury2024}
Sifat~Ahmed Chowdhury et~al.
\newblock Harnessing large language models over transformer models for
  detecting {Bengali} depressive social media text: {A} comprehensive study.
\newblock \emph{arXiv preprint arXiv:2401.07310}, 2024.
\newblock URL \url{https://arxiv.org/abs/2401.07310}.

\bibitem[Cohen(1988)]{cohen1988}
Jacob Cohen.
\newblock \emph{Statistical Power Analysis for the Behavioral Sciences}.
\newblock Lawrence Erlbaum Associates, 2nd edition, 1988.

\bibitem[Coppersmith et~al.(2015)Coppersmith, Dredze, Harman, and
  Hollingshead]{coppersmith2015}
Glen Coppersmith, Mark Dredze, Craig Harman, and Kristy Hollingshead.
\newblock From {ADHD} to {SAD}: Analyzing the language of mental health on
  {Twitter} through self-reported diagnoses.
\newblock In \emph{Proceedings of the 2nd Workshop on Computational Linguistics
  and Clinical Psychology}. Association for Computational Linguistics, 2015.

\bibitem[Ding et~al.(2024)Ding, Zhao, Shen, Hu, Luo, Joty, and
  Si]{llmannotation2024}
Bosheng Ding, Chengwei Zhao, Tianhao Shen, Hai Hu, Yuchen Luo, Shafiq Joty, and
  Luo Si.
\newblock Is {GPT}-4 a good data annotator?
\newblock In \emph{Proceedings of the 62nd Annual Meeting of the Association
  for Computational Linguistics}. Association for Computational Linguistics,
  2024.

\bibitem[Doddapaneni et~al.(2022)Doddapaneni, Saket, Kunchukuttan, Kumar, and
  Khapra]{indicbert2022}
Sumanth Doddapaneni, Rahul Saket, Anoop Kunchukuttan, Pratyush Kumar, and
  Mitesh~M Khapra.
\newblock Towards leaving no {Indic} language behind: Building monolingual
  corpora, benchmark and models for {Indic} languages.
\newblock In \emph{Proceedings of the 60th Annual Meeting of the Association
  for Computational Linguistics}, pages 4143--4158. Association for
  Computational Linguistics, 2022.

\bibitem[Ghosh et~al.(2023)]{ghosh2023}
Tapotosh Ghosh et~al.
\newblock An attention-based hybrid architecture with explainability for
  depressive social media text detection in {Bangla}.
\newblock \emph{Expert Systems with Applications}, 2023.

\bibitem[Gururangan et~al.(2020)Gururangan, Marasovic, Swayamdipta, Lo,
  Beltagy, Downey, and Smith]{gururangan2020}
Suchin Gururangan, Ana Marasovic, Swabha Swayamdipta, Kyle Lo, Iz~Beltagy, Doug
  Downey, and Noah~A. Smith.
\newblock Don't stop pretraining: Adapt language models to domains and tasks.
\newblock In \emph{Proceedings of the 58th Annual Meeting of the Association
  for Computational Linguistics}, pages 8342--8360. Association for
  Computational Linguistics, 2020.
\newblock URL \url{https://aclanthology.org/2020.acl-main.740}.

\bibitem[Islam et~al.(2020)Islam, Jubayer, Ahmed, et~al.]{islam2020}
Md~Saiful Islam, Faisal~Md Jubayer, Md~Iftekharul Ahmed, et~al.
\newblock A study on the effect of pre-processing on {Bengali} sentiment
  analysis.
\newblock In \emph{Proceedings of International Joint Conference on
  Computational Intelligence}. Springer, 2020.

\bibitem[Islam et~al.(2021)]{islam2021}
Md~Saiful Islam et~al.
\newblock A comprehensive survey on {Bengali} natural language processing.
\newblock \emph{arXiv preprint}, 2021.

\bibitem[Ji et~al.(2022)Ji, Zhang, Ansari, Fu, Martins, and Cambria]{ji2022}
Shaoxiong Ji, Tianlin Zhang, Luna Ansari, Jie Fu, Prayag Martins, and Erik
  Cambria.
\newblock {MentalBERT}: Publicly available pretrained language models for
  mental healthcare.
\newblock In \emph{Proceedings of the Language Resources and Evaluation
  Conference ({LREC} 2022)}, pages 7184--7190. European Language Resources
  Association, 2022.
\newblock URL \url{https://aclanthology.org/2022.lrec-1.778}.

\bibitem[Khanuja et~al.(2021)Khanuja, Bansal, Mehtani, Khosla, Dalmia, Sidhant,
  Maheshwari, and Talukdar]{khanuja2021}
Simran Khanuja, Diksha Bansal, Sarvesh Mehtani, Savya Khosla, Atreyee Dalmia,
  Priyanshi Sidhant, Akshit Maheshwari, and Partha Talukdar.
\newblock {MuRIL}: Multilingual representations for {Indian} languages.
\newblock \emph{arXiv preprint arXiv:2103.10730}, 2021.
\newblock URL \url{https://arxiv.org/abs/2103.10730}.

\bibitem[Matthews(1975)]{matthews1975}
Brian~W. Matthews.
\newblock Comparison of the predicted and observed secondary structure of {T4}
  phage lysozyme.
\newblock \emph{Biochimica et Biophysica Acta}, 405\penalty0 (2):\penalty0
  442--451, 1975.

\bibitem[McNemar(1947)]{mcnemar1947}
Quinn McNemar.
\newblock Note on the sampling error of the difference between correlated
  proportions or percentages.
\newblock \emph{Psychometrika}, 12\penalty0 (2):\penalty0 153--157, 1947.

\bibitem[Mumu et~al.(2023)Mumu, Tanchi, Tabassum, and Nayan]{mumu2023}
Jannatul~Ferdaous Mumu, Khushi~Noor Tanchi, Tasmiah Tabassum, and Ahsanul~Islam
  Nayan.
\newblock Depression detection from {Bangla} social media posts using
  transformers.
\newblock In \emph{Proceedings of the International Conference on Natural
  Language Processing and Machine Learning}, 2023.

\bibitem[Neuropark et~al.(2021)Neuropark, {Yandex}, and
  {HuggingFace}]{sahaj2021}
Neuropark, {Yandex}, and {HuggingFace}.
\newblock {SahajBERT}: Collaborative bengali language model via distributed
  training.
\newblock HuggingFace Model Hub, 2021.
\newblock URL \url{https://huggingface.co/neuropark/sahajBERT}.

\bibitem[{Qwen Team}(2025)]{qwen2025}
{Qwen Team}.
\newblock {Qwen2.5}: A party of foundation language models.
\newblock Hugging Face, 2025.
\newblock URL \url{https://huggingface.co/Qwen/Qwen2.5-7B}.

\bibitem[Romim et~al.(2021)Romim, Ahmed, Laskar, and Islam]{banglahate}
Nauros Romim, Mohsinul~Kabir Ahmed, Md~Tahmid~Rahman Laskar, and Md~Saiful
  Islam.
\newblock Hate speech detection in the {Bengali} language: A dataset and its
  baseline evaluation.
\newblock In \emph{Proceedings of International Conference on Big Data, IoT,
  and Machine Learning}. Springer, 2021.

\bibitem[Roy et~al.(2025)Roy, Rahaman, Shibly, Joy, Al~Kafi, and
  Sadeque]{roy2025}
Ayon Roy, Risat Rahaman, Sadat Shibly, Udoy~Saha Joy, Abdulla Al~Kafi, and
  Farig~Yousuf Sadeque.
\newblock Advancing {Bangla} machine translation through informal datasets.
\newblock \emph{arXiv preprint arXiv:2512.13487}, 2025.
\newblock URL \url{https://arxiv.org/abs/2512.13487}.

\bibitem[Sazzed(2021)]{sentnob}
Salim Sazzed.
\newblock {SentNoB}: A dataset for analysing sentiment on noisy {Bangla} texts.
\newblock Hugging Face Datasets, 2021.
\newblock URL \url{https://huggingface.co/datasets/Sazzed/sentnob}.

\bibitem[Uddin et~al.(2022)]{uddin2022}
Md~Rafiqul~Islam Uddin et~al.
\newblock Detecting depression from {Bengali} social media posts using deep
  learning.
\newblock \emph{Journal of Healthcare Engineering}, 2022.

\bibitem[Vos et~al.(2020)]{bangladesh2022}
Theo Vos et~al.
\newblock Global burden of 369 diseases and injuries in 204 countries and
  territories, 1990--2019.
\newblock \emph{The Lancet}, 396:\penalty0 1204--1222, 2020.

\bibitem[{World Health Organization}(2022)]{who2023}
{World Health Organization}.
\newblock World mental health report: Transforming mental health for all.
\newblock Technical report, WHO, 2022.
\newblock URL \url{https://www.who.int/publications/i/item/ 9789240049338}.

\bibitem[Zhai et~al.(2024)]{zhai2024}
Wei Zhai et~al.
\newblock {Chinese-MentalBERT}: Domain-adaptive pre-training on social media
  for {Chinese} mental health text analysis.
\newblock \emph{arXiv preprint arXiv:2402.09151}, 2024.
\newblock URL \url{https://arxiv.org/abs/2402.09151}.

\end{thebibliography}

\end{document}